\documentclass{article} 
\usepackage{collas2024_conference,times}
\usepackage{easyReview}

\usepackage{multirow}
\usepackage{graphicx}
\usepackage{booktabs}
\usepackage[T1]{fontenc}
\usepackage{pifont}
\newcommand{\cmark}{\ding{51}}%
\newcommand{\xmark}{\ding{55}}%

\makeatletter
\newcommand{\printfnsymbol}[1]{%
  \textsuperscript{\@fnsymbol{#1}}%
}

\usepackage{amsmath,amsfonts,bm}

\def\eqref#1{equation~\ref{#1}}

\def\1{\bm{1}}

\DeclareMathAlphabet{\mathsfit}{\encodingdefault}{\sfdefault}{m}{sl}
\SetMathAlphabet{\mathsfit}{bold}{\encodingdefault}{\sfdefault}{bx}{n}

\usepackage{hyperref}
\hypersetup{
    colorlinks=true,
    linkcolor=red,
    filecolor=magenta,
    urlcolor=blue,
    citecolor=purple,
    pdftitle={Overleaf Example},
    pdfpagemode=FullScreen,
    }

\title{Beyond Unimodal Learning: \\The Importance of Integrating Multiple Modalities for Lifelong Learning}

\author{Fahad Sarfraz\textsuperscript{1}, Bahram Zonooz\textsuperscript{1,}\thanks{Equal advisory role.} , Elahe Arani\textsuperscript{1,2,}\printfnsymbol{1}\\
\textsuperscript{1} Dep. of Mathematics and Computer Science, Eindhoven University of Technology (TU/e), Netherlands\\
\textsuperscript{2} Wayve Technologies Ltd, London, United Kingdom\\
\texttt{\{f.sarfraz, b.zonooz, e.arani\}@tue.nl}\\
}

\collasfinalcopy 

\begin{document}

\maketitle

\begin{abstract}
While humans excel at continual learning (CL), deep neural networks (DNNs) exhibit catastrophic forgetting. A salient feature of the brain that allows effective CL is that it utilizes multiple modalities for learning and inference, which is underexplored in DNNs. Therefore, we study the role and interactions of multiple modalities in mitigating forgetting and introduce a benchmark for multimodal continual learning. Our findings demonstrate that leveraging multiple views and complementary information from multiple modalities enables the model to learn more accurate and robust representations. This makes the model less vulnerable to modality-specific regularities and considerably mitigates forgetting. Furthermore, we observe that individual modalities exhibit varying degrees of robustness to distribution shift. Finally, we propose a method for integrating and aligning the information from different modalities by utilizing the relational structural similarities between the data points in each modality. Our method sets a strong baseline that enables both single- and multimodal inference. Our study provides a promising case for further exploring the role of multiple modalities in enabling CL and provides a standard benchmark for future research.\footnote{Code is publicly available at \url{https://github.com/NeurAI-Lab/MultiModal-CL}.}
\end{abstract}


\section{Introduction}

Lifelong learning requires the learning agent to continuously adapt to new data while retaining and consolidating previously learned knowledge. This ability is essential for the deployment of deep neural networks (DNNs) in numerous real-world applications. However, one critical issue in enabling continual learning (CL) in DNNs is catastrophic forgetting, whereby the model drastically forgets previously acquired knowledge when required to learn new tasks in sequence~\citep {mccloskey1989catastrophic}. Overcoming catastrophic forgetting is essential to enable lifelong learning in DNNs and make them suitable for deployment in dynamic and evolving environments. 

On the other hand, the human brain excels at CL. A salient feature of the brain that may play a critical role in enhancing its lifelong learning capabilities is that it processes and integrates information from multiple modalities. Various studies have shown that sensory modalities are integrated to facilitate perception and cognition instead of processing them independently~\citep{mroczko2016perception}. In particular, integrating audio and visual information has been shown to lead to a more accurate representation of the environment, which improves perceptual learning and memory~\citep{mcdonald2000involuntary}. Therefore, the multimodal approach to learning enhances the brain's ability to acquire and consolidate new information. 

We hypothesize that integrating multimodal learning into DNNs can similarly enhance their lifelong learning capability. By combining information from different modalities, the models can develop a more comprehensive understanding of the environment as it receives multiple views of the object, leading to a more accurate and robust representation, which is less sensitive to modality-specific regularities. In recent years, there have been several studies on how to optimally combine multiple modalities, such as vision, audio, and text, and multimodal learning has shown promise in various computer vision applications, including image captioning~\citep{wu2016value} and object recognition~\citep{karpathy2014large}. However, the efficacy of multiple modalities in continual learning and an optimal method for integrating them to mitigate forgetting is understudied, particularly in challenging scenarios such as class-incremental and domain-incremental learning.

\begin{figure}[t]
    \centering
    \includegraphics[width=\linewidth]{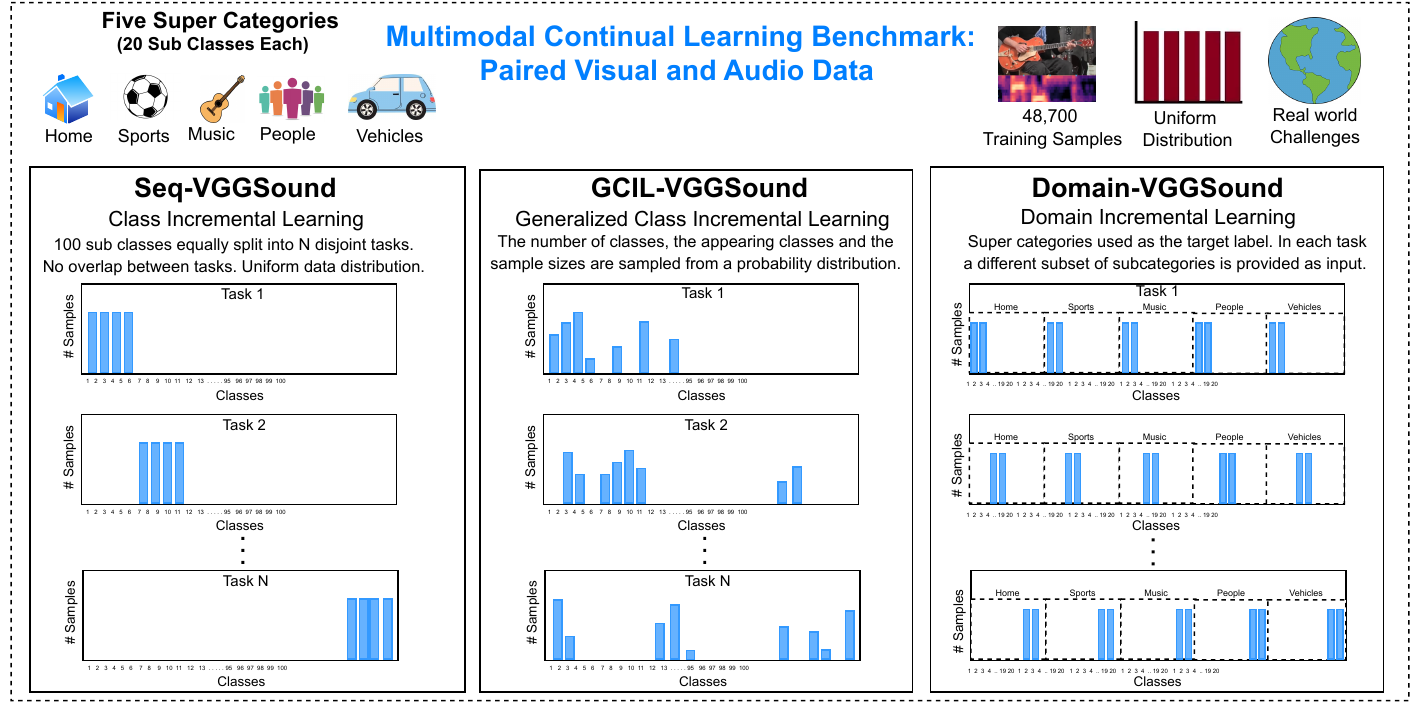}
    \caption{Multimodal continual learning (MMCL) benchmark is based on a subset of the VGGSound dataset (for increased accessibility to the research community), which covers five super categories with twenty sub-classes each. The three CL scenarios cover various challenges that a learning agent has to tackle in the real world and provide correspondence with unimodal benchmarks.}
    \label{fig:mm-benchmark}
\end{figure}

Our study addresses this research gap by studying the role and interactions of multiple modalities in mitigating forgetting in challenging and realistic continual learning settings. Our analysis demonstrates that learning from multiple modalities allows the model to learn more robust and general representations, which are less prone to forgetting and generalize better across tasks compared to learning from single modalities. Furthermore, we show that multimodal learning provides a better trade-off between the stability and plasticity of the model and reduces the bias towards recent tasks. Notably, we argue that in addition to providing complementary information about the task, different modalities exhibit different behavior and sensitivity to shifts in representations depending upon the nature of domain shift, which enhances the stability and transferability of features across the tasks. Therefore, leveraging complementary information from diverse modalities, each exhibiting varying levels of robustness to distribution shifts, can enable the model to learn a more comprehensive and robust representation of the underlying data. Improved representation facilitates better generalization and retention of knowledge across tasks, thereby enabling effective CL.

Based on the insights from our analysis, we propose a rehearsal-based multimodal CL method that utilizes the relational structural similarities between data points in each modality to integrate and align information from different modalities. Our approach allows the model to learn modality-specific representations from the visual and audio domains, which are then aligned and consolidated in such a manner that a similar relational structure of the data points is maintained in the modality-specific representations as well as the fused representation space. This facilitates the integration and alignment of the two modalities and the subsequent consolidation of knowledge across tasks. Furthermore, our method enables improved inference with both single- and multiple-modalities, significantly improving its applicability in real-world scenarios.

Finally, we introduce a benchmark for class- and domain-incremental learning based on the VGGSound dataset \citep{chen2020vggsound}, which provides a standardized evaluation platform for the community to compare and develop methods for multimodal incremental learning. Our benchmark focuses on challenging scenarios where data is received incrementally over time, simulating real-world scenarios where models need to adapt to new classes or domains without forgetting previously learned information and are exposed to additional challenges including class imbalance, learning over multiple recurrences of objects across tasks, and non-uniform distributions of classes over tasks. We empirically demonstrate the effectiveness of our approach on this benchmark and provide a strong baseline for future work. Overall, our study presents a compelling case for further exploring multimodal continual learning and provides a platform for the development and benchmarking of more robust and efficient multimodal lifelong learning methods.

\subsection{Related Work} \label{related}
The various approaches to addressing catastrophic forgetting in continual learning (CL) can be categorized into three main groups. \textit{Regularization-based} methods~\citep{kirkpatrick2017overcoming, ritter2018online, zenke2017continual,li2017learning} involve applying regularization techniques to penalize changes in the parameter or functional space of the model. \textit{Dynamic architecture} methods~\citep{yoon2018lifelong,bhat2022task} expand the network to allocate separate parameters for each task. \textit{Rehearsal-based} methods~\citep{riemer2018learning,arani2021learning,buzzega2020dark} mitigate forgetting by maintaining an episodic memory buffer and rehearsing samples from previous tasks. Among these, rehearsal-based methods have proven to be effective in challenging CL scenarios. A recent approach by \cite{vijayan2024trire} integrates all three mechanisms, offering a comprehensive solution to catastrophic forgetting. However, CL has been predominantly studied in an unimodal setting in the visual domain, and the effect and role of multiple modalities in mitigating forgetting in CL are understudied. Note that the majority of unimodal methods cannot be directly applied in multimodal settings and require considerable adaptation to the multimodal setting to make it work and further considerations to optimally utilize the complementary knowledge in multiple modalities.

However, there has been substantial progress in learning from multiple modalities~\citep{bayoudh2021survey}. Various research directions have been explored, depending on specific applications. Some studies have focused on exploring the unsupervised correspondence between multimodal data to learn meaningful representations for downstream tasks~\citep{alwassel2020self,hu2019deep,hu2019dense}. Extensive research is dedicated to leveraging information from multiple modalities to improve performance in specific tasks such as action recognition~\citep{gao2020listen,kazakos2019epic}, audio-visual speech recognition~\citep{hu2016temporal,potamianos2004audio}, visual question answering~\citep{wu2017visual,ilievski2017multimodal} and object recognition~\citep{peng2022balanced}. Furthermore, several studies~\citep{blakeman2020complementary,zhang2022self} show that complementary information enables the model to learn more holistic representations, leading to better performance. However, these studies focus on generalization in the i.i.d. setting, and progress has not been sufficiently transferred to the CL setting. A recent study~\citep{srinivasan2022climb} provides a benchmark for multimodal CL; however, they focus on vision and language tasks where each task differs significantly from the other, and their setting does not adhere to the key desiderata outlined in \cite{farquhar2018towards}. Importantly, it lacks correspondence with the challenging and established unimodal CL scenarios~\citep{van2019three}, which is essential to assess the benefits of multimodal learning compared to unimodal learning and leverages progress in the unimodal CL literature. Our study aims to fill this gap. 

\section{Multimodal Continual Learning Benchmark}
To fully explore the potential and benefits of the increasing amount of multimodal data in real-world applications in enhancing the lifelong learning capability of DNNs, it is imperative to extend the traditional unimodal CL benchmarks to encompass multimodal scenarios.
Therefore, we introduce a standardized \textit{Multimodal Continual Learning} (MMCL) benchmark (Figure \ref{fig:mm-benchmark}), which aims to simulate challenging and realistic real-world CL scenarios while maintaining correspondence with unimodal CL benchmarks and scenarios~\citep{van2019three}.

\begin{figure}[t]
    \centering
    \includegraphics[width=1\linewidth]{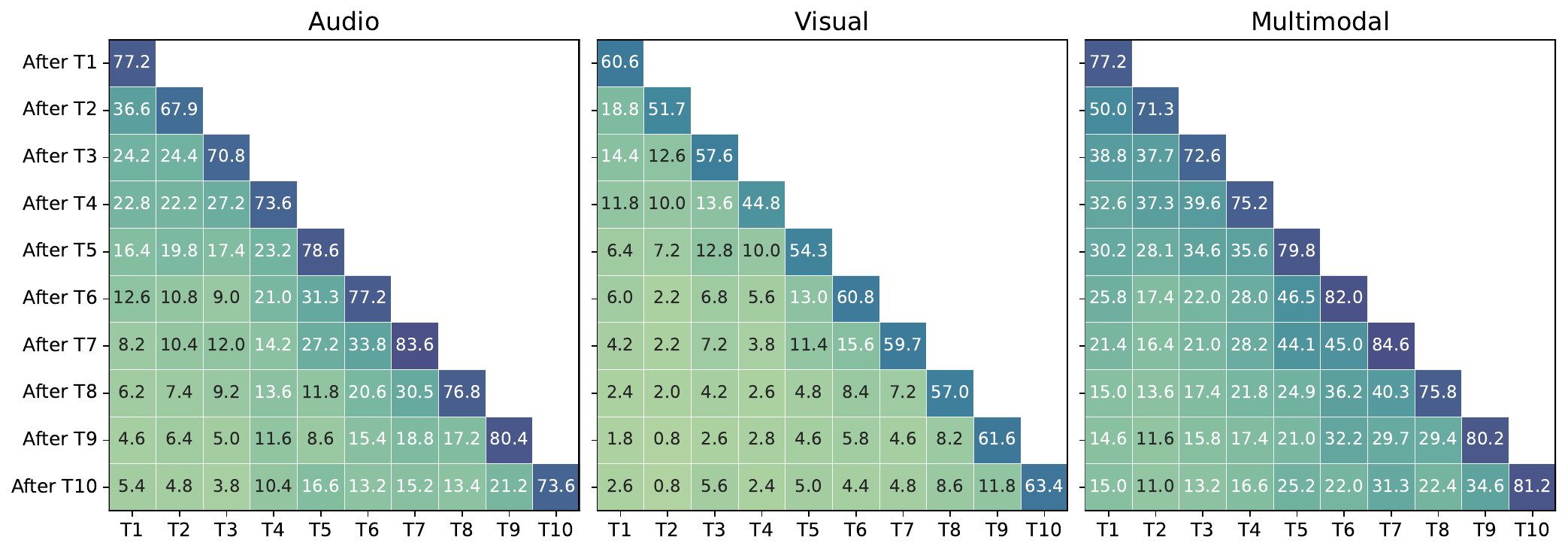}
    \caption{Taskwise performance of models trained with experience replay (1000 buffer size) on multimodal vs. unimodal (audio and visual) data on Seq-VGGSound. As we train on new tasks, T (y-axis), we monitor the performance on earlier trained tasks (x-axis). Multimodal training not only learns the new task better but also retains more performance of earlier tasks.}
    \label{fig:taskwise}
\end{figure}


MMCL benchmark is built upon the VGGSound dataset~\citep{chen2020vggsound}, which provides a diverse collection of corresponding audio and visual cues associated with challenging objects and actions in the real world and allows the exploration of multimodal learning scenarios. To ensure accessibility to a wider research community, we select a more uniform subset from the VGGSound dataset, with a total number of samples similar to CIFAR datsets~\citep{krizhevsky2009learning} ($\sim$50000 samples uniformly distributed across 100 classes), thus mitigating the requirement for extensive computational resources and memory. We present three distinct CL scenarios within the MMCL benchmark.

\textbf{Seq-VGGSound:} This scenario simulates the setting of Class-Incremental Learning (Class-IL), where a subset of 100 classes is uniformly divided into a disjoint set of N tasks. As new classes are introduced in each subsequent task, the learning agent must differentiate not only between classes within the current task, but also between classes encountered in earlier tasks. we randomly shuffled the classes and divided the dataset into 10 disjoint tasks, each containing 10 classes. 
Class-IL evaluates the method's ability to accumulate and consolidate knowledge and transfer acquired knowledge to efficiently learn generalized representations and decision boundaries for all encountered classes.

\textbf{Dom-VGGSound:} This scenario simulates the Domain-Incremental Learning (Domain-IL) setting, where the input distribution changes while the output distribution remains the same. To achieve this, we consider the supercategory of classes as the target label, and in each subsequent task, we introduce a new set of sub-classes. We consider five supercategories (Home, Sports, Music, People, and Vehicles). Domain-IL assesses the agent's capability to learn generalized features that are robust to changes in input distribution and to transfer knowledge across different domains.

\textbf{GCIL-VGGSound:} This scenario simulates the Generalized Class Incremental Learning setting (GCIL), which captures additional challenges encountered in real-world scenarios where task boundaries are blurry. The learning agent must learn from a continuous stream of data, where classes can reappear and have varying data distributions. It employs probabilistic modeling to randomly select classes and distributions. The quantity of classes per task is not predetermined, and there can be overlaps between classes, with varying sample sizes for each class. In addition to preventing catastrophic forgetting, the CL method must address sample efficiency, imbalanced classes, and efficient knowledge transfer. GCIL-VGGSound introduces assimilated challenges to evaluate the robustness and adaptability of the learning method. For further details, please refer to Appendix.

\subsection{Experimental Setup} 
\label{exp_setup}
For all our multimodal experiments, we follow the setup in \cite{peng2022balanced} and employ ResNet18~\citep{he2016deep} architecture as the backbone. The visual encoder takes multiple frames as input, following~\cite{peng2022balanced,zhao2018sound}, while the audio encoder modifies the input channel of ResNet18 from 3 to 1, as done in~\cite{chen2020vggsound}. The videos in the VGGSound dataset have a duration of 10 seconds, and we extract frames at a rate of 1 frame per second and uniformly sample 4 frames from each 10-second video clip as visual inputs in the temporal order. For audio data processing, following~\cite{peng2022balanced}, we transform the entire audio into a spectrogram with dimensions of 257$\times$1,004 using the librosa~\citep{mcfee2015librosa} library. We employ a window length of 512 and an overlap of 353. To optimize our model, we use stochastic gradient descent (SGD) with a learning rate of 0.1. 
From an architecture perspective, we adapt the ResNet architecture (following~\cite{peng2022balanced} and \ref{zhao2018sound}) for the visual encoder and audio encoder. The first conv layer of both audio and visual encoder uses a kernel size of 7 and stride 2 followed by max pooling with kernel 3 and stride 2. For multimodal fusion, we use FiLM~\citep{perez2018film} to fuse the representations from audio and video encoders, which applies affine transformations to the intermediate representations to learn how to combine them effectively. The fused representations are passed to the multimodal classifier, whereas the individual representations are flattened and passed to modality-specific representations. This allows the model the flexibility to perform both unimodal and multimodal predictions. The hyperparameters for our approach are set using a small validation set. For all our experiments, we report the mean and 1 std of three different seeds.

\section{A Case for Moving beyond Unimodal Continual Learning} \label{analysis-case}

To assess the efficacy of using multiple modalities in CL, we conduct a comprehensive analysis on the Seq-VGGSound scenario, which simulates the challenging Class-IL setting. We aim to investigate the effect of integrating multiple modalities in mitigating forgetting in challenging and realistic CL settings and the different characteristics instilled in the model. To this end, we employ the experience replay (ER) method (with a 1000 buffer size) and compare its performance when learning from unimodal data (Audio and Visual) versus multimodal data. Here, our objective is to understand how multimodal learning influences the model's ability to retain previously learned knowledge while learning new tasks. 

\subsection{Improved Continual Learning Performance}
\begin{figure}[t]
    \centering
    \includegraphics[trim= 0 0 0 0.3cm, clip, width=1\linewidth]{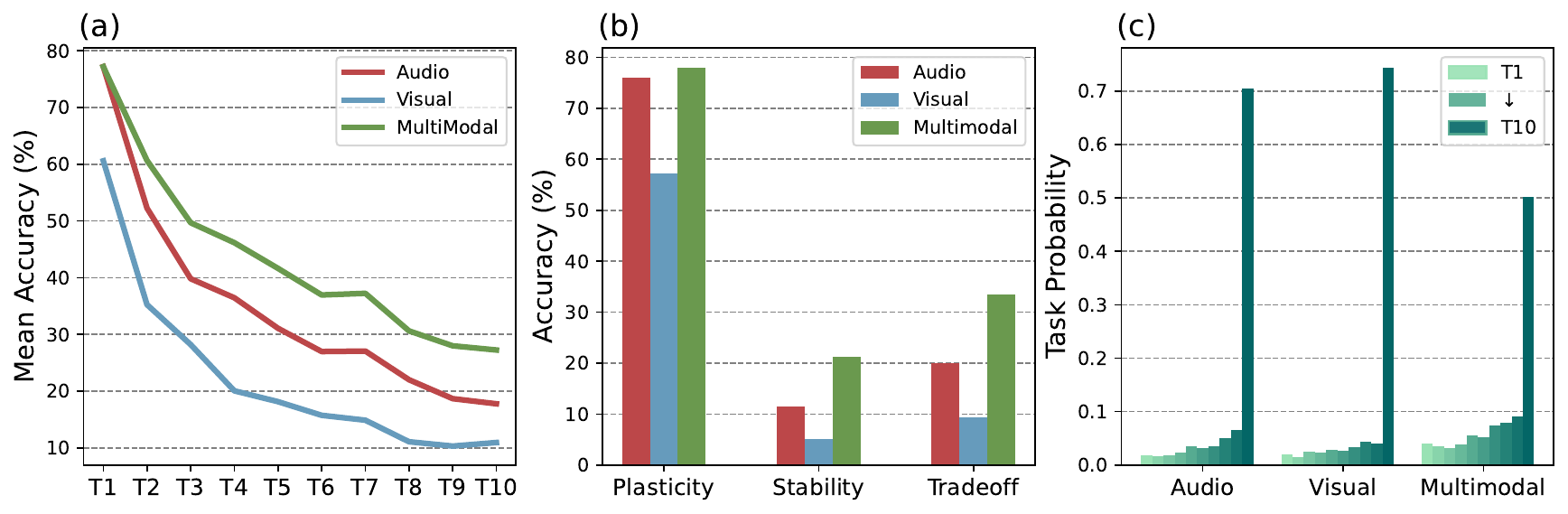}
    \caption{Comparison of experience replay with 1000 buffer size on multimodal data vs. unimodal (audio and visual) data on Seq-VGGSound. (a) shows mean accuracy on all the seen tasks (T) as training progresses. (b) provides the plasticity stability trade-off of the models while (c) compares the task recency bias. Learning with multimodal data mitigates forgetting, offers a better stability-plasticity trade-off, and reduces the bias toward recent tasks.}
    \label{fig:combined}
\end{figure}

The key challenge in CL is mitigating forgetting of earlier tasks as the model learns new tasks. A closer look at the task-wise performances of the models in Figure~\ref{fig:taskwise} shows that learning with multimodal data significantly improves both the performance of the model on the current task and the performance retention of previous tasks compared to learning from single modalities. This is further demonstrated in the mean accuracy of the models in Figure~\ref{fig:combined}(a). Interestingly, we observe that the different modalities exhibited varying levels of generalizability and susceptibility to forgetting. Notably, the audio domain provides significantly better performance and lower levels of forgetting compared to the visual domain. 
We argue that different modalities might exhibit different behavior and sensitivity in terms of shifts in representations depending upon the nature of domain shift. For instance, moving from daylight scenarios to nighttime would incur a greater shift in the visual domain compared to audio, whereas moving from an indoor to an outdoor setting may incur a higher shift in the audio domain. Therefore, leveraging the complementary information from different modalities, which shows different degrees of robustness to distribution shifts, can enable the models to capture a more comprehensive and robust representation of the underlying data, leading to improved generalization and retention of knowledge across tasks. 

\subsection{Stability Plasticity Trade-off}
To further investigate the efficacy of multimodal learning in addressing the stability-plasticity dilemma that is central to CL, we follow the analysis in~\cite{sarfraz2022synergy} to quantify the trade-off between stability and plasticity of models trained with different modalities. Given the task performance matrix $\mathcal{T}$ (Figure~\ref{fig:taskwise}), stability ($S$) is defined as the average performance of all previous $t-1$ tasks after the learning task $t$, while plasticity ($P$) is quantified by the average performance of the tasks when they are initially learned, computed as \textit{mean(Diag($\mathcal{T}$))}. Finally, the trade-off is given by $(2\times S\times P)/(S+P)$. 

Stability reflects the model's ability to avoid forgetting and maintain performance on previous tasks, while plasticity captures the model's capacity to learn new tasks. Achieving effective CL requires finding an optimal balance between stability and plasticity. Figure~\ref{fig:combined}(b) shows that multimodal learning considerably improves both the plasticity and stability of the model compared to unimodal training and thus a much better trade-off. 

\subsection{Task Recency Bias}
The sequential learning process in CL introduces a bias in the model towards the most recent task, significantly affecting its performance on previous tasks~\citep{wu2019large}. To investigate the potential of multimodal learning in mitigating the recency bias, we assess the probability of predicting each task at the end of training. To measure the probability, we utilize the softmax output of each sample in the test set and calculate the average probabilities of the classes associated with each task. Figure~\ref{fig:combined}(c) shows that multimodal learning considerably reduces the bias towards the recent task compared to learning from unimodal data. This provides valuable insight into the effectiveness of multimodal learning in addressing performance degradation in earlier tasks and highlights its potential to mitigate the bias inherent in sequential task learning in CL.

Overall, our analysis provides a compelling case for leveraging multiple modalities to mitigate forgetting in CL. The observations of improved CL performance, better plasticity stability trade-off, and reduced recency bias combined with the insights on complementary information and the differential impact of task transitions across modalities underscore the importance of multimodal integration and alignment for effective CL in DNNs. 

\section{Structure-aware Multimodal Continual Learning}
Building upon the insights from our analysis, we present a novel rehearsal-based \textit{semantic aware multimodal} continual learning method, called SAMM, which leverages the relational structure between data points in individual modalities to facilitate the consolidation of individual representations into the combined multimodal representation space. Our approach aims to integrate and align information from different modalities while maintaining a similar relational structure across the modality-specific and fused representation spaces. The goal is to allow the model to learn modality-specific representations from the visual and audio domains while utilizing the inherent structural similarities between data points to align the representations of respective modalities. In the consolidation step, the fused representation space is formed by combining the aligned modality-specific representations. This consolidated representation captures the complementary information from both modalities, resulting in a more comprehensive and robust representation of the underlying data. This enhances knowledge transfer and retention, enabling the model to leverage the combined knowledge learned across tasks. One notable advantage of our method is its versatility and applicability to both single and multiple modalities. By improving the performance of individual modalities and facilitating their integration, our approach can effectively harness the benefits of multimodal learning while accommodating scenarios where only one modality is available or applicable. 
This flexibility enhances the utility of our method in real-world scenarios where multimodal data may not always be available or one source is noisy or corrupt.

\begin{figure}[t]
    \centering
    \includegraphics[trim=1.6cm 0 0.5cm 0, clip, width=.9\linewidth]{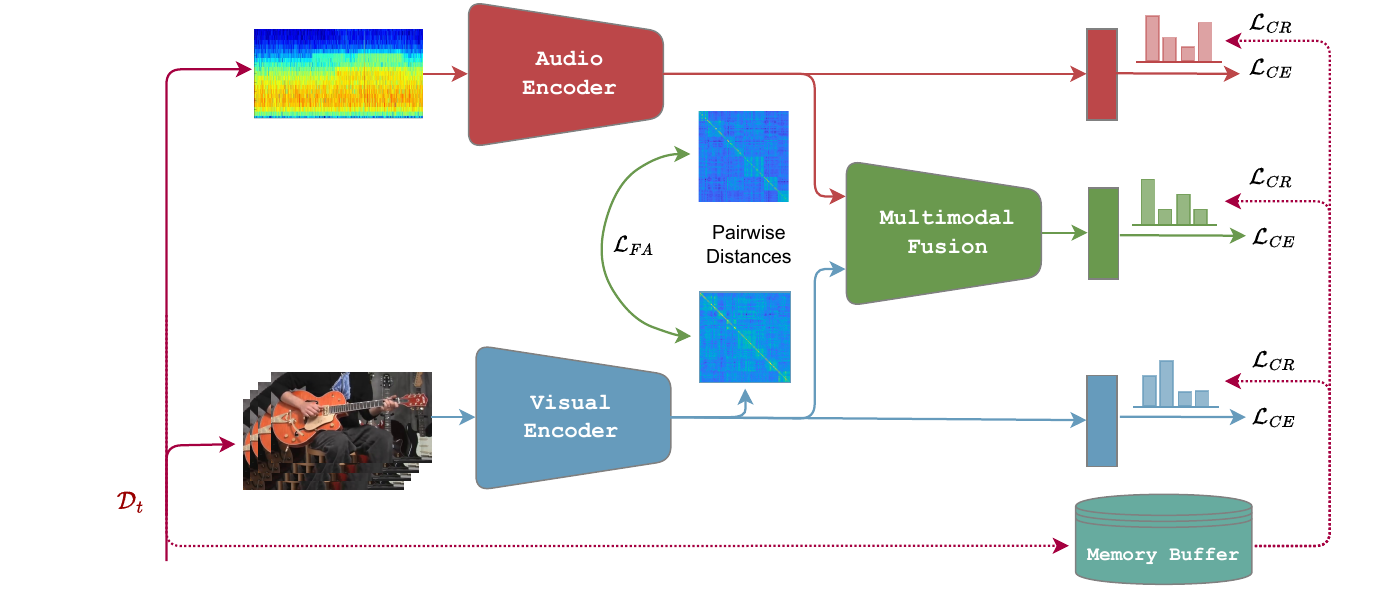}
    \caption{Semantic-Aware Multimodal (SAMM) approach leverages the relational structural information between data points in each modality to integrate and align information from different modalities. Modality-specific representations are learned in the visual and audio encoders, which are then aligned and consolidated to maintain a similar relational structure across the modality-specific representations and the fused representation space. This integration and alignment process enables the consolidation of knowledge across tasks. 
    }
    \label{fig:samm}
\end{figure}

\subsection{Components}
Our approach involves training a unified multimodal architecture on a continuous video (paired audio and visual data) stream, $\mathcal{D}$ containing a sequence of $T$ tasks ($\mathcal{D}_1$, $\mathcal{D}_2$, ..., $\mathcal{D}_T$). The model comprises an audio encoder $f(.;\theta_a)$ and a visual encoder $f(.;\theta_v)$ followed by a fusion encoder which fuses the audio and visual representations to form the multimodal representations $f(.;\theta_{av})$. Finally, the architecture consists of classification heads for the single modalities ($g(.;\phi_a)$ for audio and $g(.;\phi_v)$ for visual) and multimodal representations ($g(.;\phi_{av})$). For brevity, we represent the audio and visual features as $f_a$ = $f(x_a;\theta_{a})$ and $f_v$ = $f(x_v;\theta_{v})$ respectively, and the subsequent multimodal features as $f_{av}$ = $f(f_a, f_v;\theta_{av})$. Superscripts $t$ and $b$ indicate features corresponding to task and buffer samples, respectively.

\subsubsection{Multimodal and Unimodal Task Loss}
In addition to the supervised loss on the multimodal classifier, we also train the individual modality classification heads. The benefits are twofold: first, it allows the model to make unimodal inferences, which substantially enhances its applicability in scenarios where either only a single modality is available, or the signal from one modality is corrupted; second, it encourages the task-specific modality to learn richer features and increases the alignment between the feature space of individual modalities, which facilitates the subsequent consolidation in the common multimodal representation space. The supervision loss on the task samples $(x_a^t, x_v^t, y^t) \sim \mathcal{D}_t$ is given by:
\begin{equation} \label{eq:task_loss}
    \mathcal{L}_{s}^t =\mathcal{L}_{CE}(g(f_{av}^t; \phi_{av}), y^t) + \lambda (\mathcal{L}_{CE}(g(f_a^t; \phi_{a}), y^t) + \mathcal{L}_{CE}(g(f_v^t; \phi_{v}), y^t))
\end{equation}
where $\lambda$ (set to 0.01 for all experiments) is the regularization weight that controls the relative weightage between unimodal and multimodal performance. $\mathcal{L}_{CE}$ refers to cross-entropy loss.


\subsubsection{Experience Replay}
A common and effective approach in CL to mitigate catastrophic forgetting is the replay of samples from previous tasks stored in a small episodic memory buffer $\mathcal{M}$. To maintain the buffer, we employ reservoir sampling~\citep{vitter1985random}, which attempts to match the distribution of the data stream by ensuring that each sample in the data stream has an equal probability of being represented in the buffer and randomly replaces existing memory samples. At any given time, the distribution of the samples in the buffer approximates the distribution of all the samples encountered so far. 
At each training step, we sample a random batch $(x_a^b, x_v^b, y^b) \sim \mathcal{M}$ from the buffer and apply the supervised classification loss:
\begin{equation} \label{eq:buff_loss}
    \mathcal{L}_{s}^b =\mathcal{L}_{CE}(g(f_{av}^b; \phi_{av}), y^b) + \lambda (\mathcal{L}_{CE}(g(f_a^b; \phi_{a}), y^b) + \mathcal{L}_{CE}(g(f_v^b; \phi_{v}), y^b))
\end{equation}

\subsubsection{Consistency Regularization}
While the replay of samples from previous tasks can help alleviate catastrophic forgetting, it struggles to accurately approximate the joint distribution of all tasks encountered thus far, particularly when using smaller buffer sizes. To address this limitation, additional information from the model's earlier state is necessary to maintain the parameters closer to their optimal minima for previous tasks. Similar to earlier works~\citep{buzzega2020dark,arani2021learning}, we employ consistency regularization on the model's logit response, which encodes valuable semantic information about the representations and decision boundaries, thereby facilitating knowledge retention.

In addition to storing and replaying samples from previous tasks, our approach involves saving the unimodal and multimodal output logits ($z_a$, $z_v$, $z_{av}$) of the model. The inclusion of unimodal and multimodal outputs in the consistency regularization process further strengthens the retention of semantic relations. Furthermore, to encourage consistency across the modalities and improve the quality of supervision, we employ dynamic consistency, whereby for each sample, we select reference logits $z_r$ from the modality that provides the highest softmax score for the ground truth class. By encouraging the model to maintain consistent responses across modalities, we facilitate the integration and alignment of information from different modalities, enhancing the overall performance of the model in multimodal CL scenarios. The consistency regularization loss is given by:
\begin{equation} \label{eq:cons_reg}
    \mathcal{L}_{cr}^b =\mathcal{L}_{MSE}(g(f_{av}^b; \phi_{av}), z_r^b) + \lambda (\mathcal{L}_{MSE}(g(f_a^b; \phi_{a}), z_r^b) + \mathcal{L}_{MSE}(g(f_v^b; \phi_{v}), z_r^b))
\end{equation}
where $\mathcal{L}_{MSE}$ refers to the mean squared error loss.

\subsubsection{Semantic-aware Feature Alignment}
One of the key challenges in multimodal learning is the alignment of feature representations across different modalities. This alignment is crucial for the effective consolidation of knowledge in a shared multimodal representation space. We address this challenge by using the relational structure between data points in each modality to integrate and align information from different modalities. Our method allows the model to learn modality-specific representations from both the visual and the audio domains. These modality-specific representations are then aligned and consolidated in a way that preserves a similar relational structure among the data points within each modality. This encourages the learned multimodal representations to capture the essential relationships and similarities between modalities and promotes a holistic understanding of the underlying data, improving the ability of the model to learn and transfer knowledge across tasks. 

To this end, we employ the distance-wise relation knowledge distillation loss from~\citet{park2019relational} separately on the task samples and the buffer samples:
\begin{equation} \label{eq:sim_loss}
    \mathcal{L}_{FA}^{b,t} = \sum_{(x^i, x^j) \in \chi_{b,t}^2} \mathcal{L}_H (\psi_D(f_a^i, f_a^j), \psi_D(f_v^i, f_v^j)), ~~~~~~~~~\psi_D(f^i, f^j) = \frac{1}{\mu}||f^i - f^j||_2
\end{equation}
where $\mathcal{L}_H$ is Huber loss~\citep{huber1992robust}, $\mu$ represents the mean distance between all pairs within the given batch of $\chi^2_{b,t}$, whether it is from the buffer or task samples.
Essentially, the loss encourages a similar pairwise relationship structure in the different modalities by penalizing distance differences between their individual output representation spaces.

Finally, we combine individual losses to train the multimodal architecture. 
\begin{equation} \label{eq:overall_loss}
    \mathcal{L} = \mathcal{L}_{s}^t + \mathcal{L}_{s}^b + \beta \cdot \mathcal{L}_{cr}^b + (\mathcal{L}_{FA}^{t} + \mathcal{L}_{FA}^{b}) 
\end{equation}
%
where $\beta$ controls the strength of consistency regularization loss. 

\subsubsection{Dynamic Multimodal Inference}

While utilizing multiple modalities improves generalization, for each sample, ideally the model should be able to weigh each modality based on how informative it is. This allows us to deal with scenarios where one modality is corrupted, noisy, or occluded. To this end, we use a weighted ensemble of the classifiers based on the softmax confidence score:
\begin{equation} \label{eq:dynamic-inf}
    z_o = \max(\sigma(z_a)) \cdot z_a + \max(\sigma(z_v)) \cdot z_v + \max(\sigma(z_{av})) \cdot z_{av}  
\end{equation}
For such a weighting scheme to work well, the model should be well-calibrated so that the confidence score is a good proxy of the modality's performance. At the end of each task, we calibrate the classifiers using temperature scaling~\citep{guo2017calibration} on the buffer samples. This provides us with a simple and effective approach for leveraging different modalities based on their quality of signal.

\begin{table}[t]
\caption{Comparison of different methods on individual and multiple modalities on different CL scenarios based on the VGGSound dataset. We report mean and 1 s.t.d of three seeds}
\label{tab:vgg-sound}
\centering
\begin{tabular}{@{\extracolsep{4pt}}ll|ccc|ccc}
\toprule
\multirow{2}{*}{Buffer} & \multirow{2}{*}{{Method}} & \multicolumn{3}{c|}{Seq-VGGSound} & \multicolumn{3}{c}{Dom-VGGSound} \\ \cmidrule{3-8}
 &  &  Audio & Visual &  Multimodal & Audio & Visual & Multimodal\\ \midrule
 
\multirow{2}{*}{–} 
 & JOINT & 53.47\tiny±1.62  & 34.52\tiny±0.28  & 58.21\tiny±0.24  & 57.48\tiny±0.80 & 42.87\tiny±2.04 & 61.66\tiny±1.40   \\
 & SGD   & 7.60\tiny±0.33   &  6.37\tiny±0.33  &  8.24\tiny±0.09  & 26.89\tiny±0.17 & 24.80\tiny±0.12 & 27.38\tiny±0.35   \\ \midrule 
 
 \multirow{2}{*}{500} 
 & ER   & 13.92\tiny±1.07 & 9.07\tiny±0.39 & 21.44\tiny±1.76 & 31.31\tiny±0.80 & 27.36\tiny±1.60 &  31.85\tiny±1.09 \\
 & SAMM & 23.61\tiny±1.06 & 8.90\tiny±0.35 & \textbf{26.34}\tiny±0.42 & \textbf{36.27}\tiny±0.29 & 24.98\tiny±0.41 & 35.74\tiny±0.59  \\ \midrule

  \multirow{2}{*}{1000} 
 & ER   & 18.06\tiny±0.44 & 11.23\tiny±0.57 & 28.09\tiny±0.77 & 35.31\tiny±0.65 & 27.73\tiny±0.99 & 36.00\tiny±1.08 \\
 & SAMM & 28.59\tiny±0.77 & 10.08\tiny±0.34  & \textbf{34.51}\tiny±2.37 & 38.63\tiny±0.43 & 26.53\tiny±0.12 & \textbf{39.49}\tiny±0.36  \\ \midrule

  \multirow{2}{*}{2000} 
 & ER   & 23.41\tiny±0.50 & 14.19\tiny±0.32 & 34.02\tiny±0.40 & 38.23\tiny±0.72 & 29.28\tiny±0.63 &  39.30\tiny±1.55  \\
 & SAMM & 32.20\tiny±0.28 & 11.60\tiny±0.43 & \textbf{37.76}\tiny±2.94 & 42.53\tiny±0.47 & 28.12\tiny±0.31 & \textbf{43.72}\tiny±0.34  \\ \midrule
\end{tabular}
\end{table}

\section{Empirical Evaluation}
We compare our semantic-aware multimodal learning approach with the baseline Experience Replay (ER)~\citep{riemer2018learning} method under uniform experimental conditions (Section \ref{exp_setup}) on a wide range of multimodal CL scenarios that cover various challenges that a lifelong learning agent has to tackle in the real world. Note that the baseline ER method requires separate training on each individual modality and can make inferences on either multimodal data or on individual modalities. In contrast, our proposed approach utilizes a unified architecture that is able to make inferences under a multimodal setting, as well as individual modalities, enabling more efficient training and inference while enhancing the applicability of the method.

Table \ref{tab:vgg-sound} shows that SAMM improves the performance of the model in the majority of the scenarios. The improvement under Seq-VGGSound shows that SAMM effectively mitigates forgetting and can learn well-aligned general representations that are able to transfer knowledge across tasks. In particular, the considerable gains over multimodal ER show that our proposed structure-aware multimodal learning approach is able to better leverage the complementary information in different modalities to learn a more robust representation. Interestingly, we observe an order of magnitude gain in audio even compared to a model trained specifically on it. We argue that multimodal learning leads to richer feature representations that are more generalizable, and aligning the two modalities not only facilitates the consolidation in the multimodal representation space, but also allows efficient knowledge transfer between modalities. Note that visual generally performs much worse compared to audio as for many classes (e.g. clapping) audio cues are more informative and visual cues (e.g. performers on stage) can be misleading. We observe similar gains in the Domain-VGGSound setting, where the model is required to learn robust generalizable features, which are robust to input distribution. Dom-VGGSound exposes the model to a sharp input distribution shift as the subclasses change at the task transition, and hence the competitive performance of our approach in this setting suggests that it is able to learn generalizable features that are more robust to distribution shifts.

\begin{table}[t]
\caption{Performance comparison on individual and multiple modalities on GCIL-VGGSound.}
\label{tab:vgg-gcil}
\centering
\begin{tabular}{@{\extracolsep{4pt}}ll|ccc|ccc}
\toprule
\multirow{2}{*}{Buffer} & \multirow{2}{*}{{Method}} & \multicolumn{3}{c|}{Uniform} & \multicolumn{3}{c}{Longtail}\\ \cmidrule{3-8}
 &  &  Audio & Visual &  MultiModal & Audio & Visual &  MultiModal \\ \midrule
 
\multirow{2}{*}{–} 
 & JOINT & 44.02\tiny± 1.51& 26.23\tiny±0.63 & 49.32\tiny±0.43 & 43.23\tiny±1.38 & 25.19\tiny±0.76 & 47.17\tiny±0.31  \\
 & SGD   & 20.34\tiny±0.51 & 11.47\tiny±0.79 & 24.73\tiny±0.40   & 19.00\tiny±0.43 & 10.43\tiny±0.67 & 22.03\tiny±0.67  \\ \midrule
 
 \multirow{2}{*}{500} 
 & ER   & 24.57\tiny±0.44 & 13.80\tiny±0.53 & 29.76\tiny±0.82 & 24.30\tiny±0.33 & 12.81\tiny±0.11 & 28.58\tiny±0.73 \\
 & SAMM & 27.41\tiny±0.41 & 11.90\tiny±1.65 & \textbf{34.34}\tiny±0.78 & 27.25\tiny±0.65 & 12.06\tiny±0.22 & \textbf{34.16}\tiny±0.81 \\ \midrule

  \multirow{2}{*}{1000} 
 & ER    &  27.32\tiny±0.38 & 15.53\tiny±0.30 & 34.27\tiny±0.77 & 27.25\tiny±0.93 & 14.24\tiny±0.25 & 31.60\tiny±0.94 \\
 & SAMM  &  29.99\tiny±0.41 & 13.18\tiny±0.64 & \textbf{38.04}\tiny±1.51  & 28.52\tiny±0.40 & 12.64\tiny±0.12 & \textbf{36.15}\tiny±0.30 \\ \midrule

  \multirow{2}{*}{2000} 
 & ER    & 31.30\tiny±0.28 & 17.25\tiny±0.09 & 37.77\tiny±0.80 & 29.75\tiny±0.46 & 17.31\tiny±0.21 & 35.66\tiny±0.53 \\
 & SAMM  & 31.96\tiny±0.76 & 14.35\tiny±0.58 & \textbf{42.08}\tiny±1.89 & 30.13\tiny±0.68 & 13.09\tiny±0.57 & \textbf{40.33}\tiny±0.38 \\ \midrule
\end{tabular}
\end{table}

To further evaluate the versatility of our approach, we also consider GCIL, which introduces additional complexities of real-world scenarios where tasks are nonuniform, and classes can reoccur with different distributions.  Table \ref{tab:vgg-gcil} compares the performance of our method with ER under the uniform and longtail data distribution settings. Consistent gains with our method show the effectiveness of our approach in learning across multiple occurrences of the object, improving the sample efficiency, and enhancing the robustness to imbalanced data. We attribute the performance gains in our method to the efficient utilization of individual modalities by encouraging them to learn discriminative features using the modality-specific classification head and aligning them using the semantic aware feature alignment loss, which facilitates the consolidation of modalities in a common representation space. Overall, our results provide strong motivation for further exploration of multimodal learning in CL and developing methods that efficiently utilize the complementary information in different modalities and varying robustness to distribution shifts at the task transition to enable efficient CL. 

\subsection{Ablation Study}
\label{sec:abl}

In order to gain a deeper understanding of the individual contributions made by each component of the SAMM, we systematically introduced each component one at a time and evaluated its impact on the model's performance across both unimodal and multimodal inference. Table \ref{tab:ablation} shows that all components (Unimodal classifiers (UM) + Consistency Regularization (CR), Feature Alignment (FA), and Dynamic Inference (DI)) contribute to the final performance of the method.

Unimodal classifiers coupled with consistency regularization demonstrate a substantial positive impact on the performance of single modalities and also show improvements in multimodal scenarios. Furthermore, incorporation of feature alignment leads to enhanced performance in both the audio and audio-visual domains. The improvement in the multimodal scenario shows that the alignment of the individual modalities facilitates the consolidation of complementary information in the fused representation space. However, perhaps due to the nature of the datasets (where the visual domain is sometimes unrelated to the audio cues e.g. clapping of the audience in the audio domain while visual shows the performers taking a standing ovation), this alignment and fusion can lead to a decrement in the visual modality performance. Additionally, dynamic inference is specifically designed to leverage maximum knowledge from both unimodalities, and our experiments confirm its effectiveness, resulting in an approximately 10\% improvement in the performance of the multimodal system.

\begin{table}[t]
\centering
\caption{Contribution of the different components (UM: Unimodal classifiers, CR: Consistency Regularization, FA: Feature Alignment, and DI: Dynamic Inference) of SAMM on the performance of the model on Seq-VGGSound with 1000 buffer size.}
\label{tab:ablation}
\begin{tabular}{ccc|ccc}
\toprule
 $UM + CR$ & $FA$ & $DI$ & Audio & Video & Multimodal \\
 \midrule
\xmark & \xmark & \xmark & 1.05\tiny±0.03 & 1.25\tiny±0.06 & 28.09\tiny±0.77 \\
\cmark & \xmark & \xmark & 24.03\tiny±1.08 & 10.32\tiny±0.28 & 29.13\tiny±0.89 \\
\cmark & \cmark & \xmark & 28.59\tiny±0.77 & 10.08\tiny±0.34 & 31.48\tiny±1.07 \\
\cmark & \cmark & \cmark & 28.59\tiny±0.77 & 10.08\tiny±0.34 & 34.51\tiny±2.37 \\
\bottomrule
\end{tabular}
\end{table}

\begin{figure}[t]
    \centering
    \includegraphics[trim=0 0 0.5cm 0, clip, width=.73\linewidth]{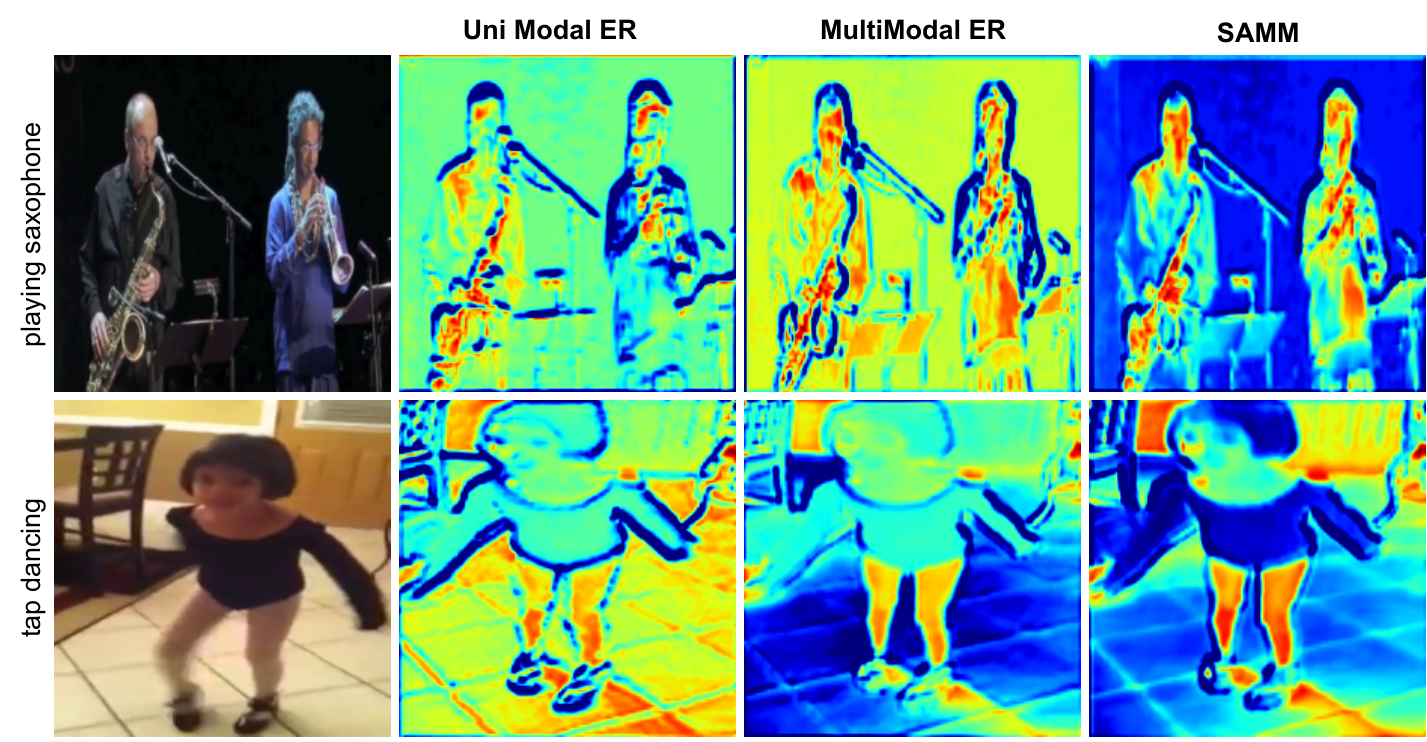}
    \caption{Gradcam activations for different models. Multimodal ER attends more to the regions that are associated with the class labels and localizes the sound regions better. This is considerably improved with SAMM.}
    \label{fig:activations}
\end{figure}

\subsection{Activation Maps}
To gain a better understanding of the effect of multimodal learning, we look at the GradCam \citep{selvaraju2017grad} activations of the model. We use the gradients of the first convolution layer of the visual net with respect to the output logit corresponding to the class label. The activation maps in Figure \ref{fig:activations} show that MultiModal ER allows the model to attend to regions of the image that are associated with the label and to localize the sound. SAMM considerably improves sound localization and attends to the most pertinent regions in the image. For playing saxophone, SAMM attends more to the saxophone and the mouth regions. Similarly, for tap dancing, SAMM rightfully attends more to the legs. This shows that multiple modalities and structure-aware alignment in SAMM enable the model to learn more holistic representations and focus on the regions associated with the class. The enhanced localization of sound in SAMM shows that our method effectively aligns the two modalities and learns a better representation.

\section{Conclusion}
Our study advocates for the integration of multiple modalities to facilitate continual learning, emphasizing the pivotal role of multisensory information processing similar to the human brain. We incorporated audio and visual modalities as video is a more natural way to represent objects and actions, similar to how humans consume information in the real world
and our empirical evaluation show reduced forgetting and enhanced adaptability, showcasing its ability to foster more robust representations of objects and actions while mitigating forgetting through multimodal replay. We also introduced a standardized Multimodal Continual Learning benchmark, tailored to simulate realistic CL scenarios and allow for direct comparison with unimodal benchmarks. Furthermore, our proposed Semantic Aware Multimodal Learning approach highlights key design tenets for effective multimodal CL. We hope that our study would encourage further exploration of this promising direction and integration of additional modalities into the framework.

\bibliography{collas2024_conference}

\begin{thebibliography}{45}
\providecommand{\natexlab}[1]{#1}
\providecommand{\url}[1]{\texttt{#1}}
\expandafter\ifx\csname urlstyle\endcsname\relax
  \providecommand{\doi}[1]{doi: #1}\else
  \providecommand{\doi}{doi: \begingroup \urlstyle{rm}\Url}\fi

\bibitem[Alwassel et~al.(2020)Alwassel, Mahajan, Korbar, Torresani, Ghanem, and
  Tran]{alwassel2020self}
Humam Alwassel, Dhruv Mahajan, Bruno Korbar, Lorenzo Torresani, Bernard Ghanem,
  and Du~Tran.
\newblock Self-supervised learning by cross-modal audio-video clustering.
\newblock \emph{Advances in Neural Information Processing Systems},
  33:\penalty0 9758--9770, 2020.

\bibitem[Arani et~al.(2022)Arani, Sarfraz, and Zonooz]{arani2021learning}
Elahe Arani, Fahad Sarfraz, and Bahram Zonooz.
\newblock Learning fast, learning slow: A general continual learning method
  based on complementary learning system.
\newblock In \emph{International Conference on Learning Representations}, 2022.
\newblock URL \url{https://openreview.net/forum?id=uxxFrDwrE7Y}.

\bibitem[Bayoudh et~al.(2021)Bayoudh, Knani, Hamdaoui, and
  Mtibaa]{bayoudh2021survey}
Khaled Bayoudh, Raja Knani, Fay{\c{c}}al Hamdaoui, and Abdellatif Mtibaa.
\newblock A survey on deep multimodal learning for computer vision: advances,
  trends, applications, and datasets.
\newblock \emph{The Visual Computer}, pp.\  1--32, 2021.

\bibitem[Bhat et~al.(2023)Bhat, Zonooz, and Arani]{bhat2022task}
Prashant~Shivaram Bhat, Bahram Zonooz, and Elahe Arani.
\newblock Task-aware information routing from common representation space in
  lifelong learning.
\newblock In \emph{The Eleventh International Conference on Learning
  Representations}, 2023.
\newblock URL \url{https://openreview.net/forum?id=-M0TNnyWFT5}.

\bibitem[Blakeman \& Mareschal(2020)Blakeman and
  Mareschal]{blakeman2020complementary}
Sam Blakeman and Denis Mareschal.
\newblock A complementary learning systems approach to temporal difference
  learning.
\newblock \emph{Neural Networks}, 122:\penalty0 218--230, 2020.

\bibitem[Buzzega et~al.(2020)Buzzega, Boschini, Porrello, Abati, and
  Calderara]{buzzega2020dark}
Pietro Buzzega, Matteo Boschini, Angelo Porrello, Davide Abati, and Simone
  Calderara.
\newblock Dark experience for general continual learning: a strong, simple
  baseline.
\newblock \emph{Advances in neural information processing systems},
  33:\penalty0 15920--15930, 2020.

\bibitem[Chen et~al.(2020)Chen, Xie, Vedaldi, and Zisserman]{chen2020vggsound}
Honglie Chen, Weidi Xie, Andrea Vedaldi, and Andrew Zisserman.
\newblock Vggsound: A large-scale audio-visual dataset.
\newblock In \emph{ICASSP 2020-2020 IEEE International Conference on Acoustics,
  Speech and Signal Processing (ICASSP)}, pp.\  721--725. IEEE, 2020.

\bibitem[Farquhar \& Gal(2018)Farquhar and Gal]{farquhar2018towards}
Sebastian Farquhar and Yarin Gal.
\newblock Towards robust evaluations of continual learning.
\newblock \emph{arXiv preprint arXiv:1805.09733}, 2018.

\bibitem[Gao et~al.(2020)Gao, Oh, Grauman, and Torresani]{gao2020listen}
Ruohan Gao, Tae-Hyun Oh, Kristen Grauman, and Lorenzo Torresani.
\newblock Listen to look: Action recognition by previewing audio.
\newblock In \emph{Proceedings of the IEEE/CVF Conference on Computer Vision
  and Pattern Recognition}, pp.\  10457--10467, 2020.

\bibitem[Guo et~al.(2017)Guo, Pleiss, Sun, and Weinberger]{guo2017calibration}
Chuan Guo, Geoff Pleiss, Yu~Sun, and Kilian~Q Weinberger.
\newblock On calibration of modern neural networks.
\newblock In \emph{International Conference on Machine Learning}, pp.\
  1321--1330. PMLR, 2017.

\bibitem[He et~al.(2016)He, Zhang, Ren, and Sun]{he2016deep}
Kaiming He, Xiangyu Zhang, Shaoqing Ren, and Jian Sun.
\newblock Deep residual learning for image recognition.
\newblock In \emph{Proceedings of the IEEE conference on computer vision and
  pattern recognition}, pp.\  770--778, 2016.

\bibitem[Hu et~al.(2016)Hu, Li, et~al.]{hu2016temporal}
Di~Hu, Xuelong Li, et~al.
\newblock Temporal multimodal learning in audiovisual speech recognition.
\newblock In \emph{Proceedings of the IEEE Conference on Computer Vision and
  Pattern Recognition}, pp.\  3574--3582, 2016.

\bibitem[Hu et~al.(2019{\natexlab{a}})Hu, Nie, and Li]{hu2019deep}
Di~Hu, Feiping Nie, and Xuelong Li.
\newblock Deep multimodal clustering for unsupervised audiovisual learning.
\newblock In \emph{Proceedings of the IEEE/CVF Conference on Computer Vision
  and Pattern Recognition}, pp.\  9248--9257, 2019{\natexlab{a}}.

\bibitem[Hu et~al.(2019{\natexlab{b}})Hu, Wang, Nie, and Li]{hu2019dense}
Di~Hu, Chengze Wang, Feiping Nie, and Xuelong Li.
\newblock Dense multimodal fusion for hierarchically joint representation.
\newblock In \emph{ICASSP 2019-2019 IEEE International Conference on Acoustics,
  Speech and Signal Processing (ICASSP)}, pp.\  3941--3945. IEEE,
  2019{\natexlab{b}}.

\bibitem[Huber(1992)]{huber1992robust}
Peter~J Huber.
\newblock Robust estimation of a location parameter.
\newblock In \emph{Breakthroughs in statistics: Methodology and distribution},
  pp.\  492--518. Springer, 1992.

\bibitem[Ilievski \& Feng(2017)Ilievski and Feng]{ilievski2017multimodal}
Ilija Ilievski and Jiashi Feng.
\newblock Multimodal learning and reasoning for visual question answering.
\newblock \emph{Advances in neural information processing systems}, 30, 2017.

\bibitem[Karpathy et~al.(2014)Karpathy, Toderici, Shetty, Leung, Sukthankar,
  and Fei-Fei]{karpathy2014large}
Andrej Karpathy, George Toderici, Sanketh Shetty, Thomas Leung, Rahul
  Sukthankar, and Li~Fei-Fei.
\newblock Large-scale video classification with convolutional neural networks.
\newblock In \emph{Proceedings of the IEEE conference on Computer Vision and
  Pattern Recognition}, pp.\  1725--1732, 2014.

\bibitem[Kazakos et~al.(2019)Kazakos, Nagrani, Zisserman, and
  Damen]{kazakos2019epic}
Evangelos Kazakos, Arsha Nagrani, Andrew Zisserman, and Dima Damen.
\newblock Epic-fusion: Audio-visual temporal binding for egocentric action
  recognition.
\newblock In \emph{Proceedings of the IEEE/CVF International Conference on
  Computer Vision}, pp.\  5492--5501, 2019.

\bibitem[Kirkpatrick et~al.(2017)Kirkpatrick, Pascanu, Rabinowitz, Veness,
  Desjardins, Rusu, Milan, Quan, Ramalho, Grabska-Barwinska,
  et~al.]{kirkpatrick2017overcoming}
James Kirkpatrick, Razvan Pascanu, Neil Rabinowitz, Joel Veness, Guillaume
  Desjardins, Andrei~A Rusu, Kieran Milan, John Quan, Tiago Ramalho, Agnieszka
  Grabska-Barwinska, et~al.
\newblock Overcoming catastrophic forgetting in neural networks.
\newblock \emph{Proceedings of the national academy of sciences}, 114\penalty0
  (13):\penalty0 3521--3526, 2017.

\bibitem[Krizhevsky et~al.(2009)]{krizhevsky2009learning}
Alex Krizhevsky et~al.
\newblock Learning multiple layers of features from tiny images.
\newblock 2009.

\bibitem[Li \& Hoiem(2017)Li and Hoiem]{li2017learning}
Zhizhong Li and Derek Hoiem.
\newblock Learning without forgetting.
\newblock \emph{IEEE transactions on pattern analysis and machine
  intelligence}, 40\penalty0 (12):\penalty0 2935--2947, 2017.

\bibitem[McCloskey \& Cohen(1989)McCloskey and
  Cohen]{mccloskey1989catastrophic}
Michael McCloskey and Neal~J Cohen.
\newblock Catastrophic interference in connectionist networks: The sequential
  learning problem.
\newblock In \emph{Psychology of learning and motivation}, volume~24, pp.\
  109--165. Elsevier, 1989.

\bibitem[McDonald et~al.(2000)McDonald, Teder-Sa{\`E}leja{\`E}rvi, and
  Hillyard]{mcdonald2000involuntary}
John~J McDonald, Wolfgang~A Teder-Sa{\`E}leja{\`E}rvi, and Steven~A Hillyard.
\newblock Involuntary orienting to sound improves visual perception.
\newblock \emph{Nature}, 407\penalty0 (6806):\penalty0 906--908, 2000.

\bibitem[McFee et~al.(2015)McFee, Raffel, Liang, Ellis, McVicar, Battenberg,
  and Nieto]{mcfee2015librosa}
Brian McFee, Colin Raffel, Dawen Liang, Daniel~P Ellis, Matt McVicar, Eric
  Battenberg, and Oriol Nieto.
\newblock librosa: Audio and music signal analysis in python.
\newblock In \emph{Proceedings of the 14th python in science conference},
  volume~8, pp.\  18--25, 2015.

\bibitem[Mi et~al.(2020)Mi, Kong, Lin, Yu, and Faltings]{mi2020generalized}
Fei Mi, Lingjing Kong, Tao Lin, Kaicheng Yu, and Boi Faltings.
\newblock Generalized class incremental learning.
\newblock In \emph{Proceedings of the IEEE/CVF Conference on Computer Vision
  and Pattern Recognition Workshops}, pp.\  240--241, 2020.

\bibitem[Mroczko-Wasowicz(2016)]{mroczko2016perception}
Aleksandra Mroczko-Wasowicz.
\newblock Perception--cognition interface and cross-modal experiences: Insights
  into unified consciousness, 2016.

\bibitem[Park et~al.(2019)Park, Kim, Lu, and Cho]{park2019relational}
Wonpyo Park, Dongju Kim, Yan Lu, and Minsu Cho.
\newblock Relational knowledge distillation.
\newblock In \emph{Proceedings of the IEEE/CVF conference on computer vision
  and pattern recognition}, pp.\  3967--3976, 2019.

\bibitem[Peng et~al.(2022)Peng, Wei, Deng, Wang, and Hu]{peng2022balanced}
Xiaokang Peng, Yake Wei, Andong Deng, Dong Wang, and Di~Hu.
\newblock Balanced multimodal learning via on-the-fly gradient modulation.
\newblock In \emph{Proceedings of the IEEE/CVF Conference on Computer Vision
  and Pattern Recognition}, pp.\  8238--8247, 2022.

\bibitem[Perez et~al.(2018)Perez, Strub, De~Vries, Dumoulin, and
  Courville]{perez2018film}
Ethan Perez, Florian Strub, Harm De~Vries, Vincent Dumoulin, and Aaron
  Courville.
\newblock Film: Visual reasoning with a general conditioning layer.
\newblock In \emph{Proceedings of the AAAI conference on artificial
  intelligence}, volume~32, 2018.

\bibitem[Potamianos et~al.(2004)Potamianos, Neti, Luettin, and
  Matthews]{potamianos2004audio}
Gerasimos Potamianos, Chalapathy Neti, Juergen Luettin, and Iain Matthews.
\newblock Audio-visual automatic speech recognition: An overview.
\newblock \emph{Issues in visual and audio-visual speech processing},
  22:\penalty0 23, 2004.

\bibitem[Riemer et~al.(2019)Riemer, Cases, Ajemian, Liu, Rish, Tu, , and
  Tesauro]{riemer2018learning}
Matthew Riemer, Ignacio Cases, Robert Ajemian, Miao Liu, Irina Rish, Yuhai Tu,
  , and Gerald Tesauro.
\newblock Learning to learn without forgetting by maximizing transfer and
  minimizing interference.
\newblock In \emph{International Conference on Learning Representations}, 2019.
\newblock URL \url{https://openreview.net/forum?id=B1gTShAct7}.

\bibitem[Ritter et~al.(2018)Ritter, Botev, and Barber]{ritter2018online}
Hippolyt Ritter, Aleksandar Botev, and David Barber.
\newblock Online structured laplace approximations for overcoming catastrophic
  forgetting.
\newblock In \emph{Advances in Neural Information Processing Systems}, pp.\
  3738--3748, 2018.

\bibitem[Sarfraz et~al.(2022)Sarfraz, Arani, and Zonooz]{sarfraz2022synergy}
Fahad Sarfraz, Elahe Arani, and Bahram Zonooz.
\newblock Synergy between synaptic consolidation and experience replay for
  general continual learning.
\newblock In \emph{Conference on Lifelong Learning Agents}, pp.\  920--936.
  PMLR, 2022.

\bibitem[Selvaraju et~al.(2017)Selvaraju, Cogswell, Das, Vedantam, Parikh, and
  Batra]{selvaraju2017grad}
Ramprasaath~R Selvaraju, Michael Cogswell, Abhishek Das, Ramakrishna Vedantam,
  Devi Parikh, and Dhruv Batra.
\newblock Grad-cam: Visual explanations from deep networks via gradient-based
  localization.
\newblock In \emph{Proceedings of the IEEE international conference on computer
  vision}, pp.\  618--626, 2017.

\bibitem[Srinivasan et~al.(2022)Srinivasan, Chang, Pinto~Alva, Chochlakis,
  Rostami, and Thomason]{srinivasan2022climb}
Tejas Srinivasan, Ting-Yun Chang, Leticia Pinto~Alva, Georgios Chochlakis,
  Mohammad Rostami, and Jesse Thomason.
\newblock Climb: A continual learning benchmark for vision-and-language tasks.
\newblock \emph{Advances in Neural Information Processing Systems},
  35:\penalty0 29440--29453, 2022.

\bibitem[van~de Ven \& Tolias(2019)van~de Ven and Tolias]{van2019three}
Gido~M van~de Ven and Andreas~S Tolias.
\newblock Three scenarios for continual learning.
\newblock \emph{arXiv preprint arXiv:1904.07734}, 2019.

\bibitem[Vijayan et~al.(2024)Vijayan, Bhat, Zonooz, and
  Arani]{vijayan2024trire}
Preetha Vijayan, Prashant Bhat, Bahram Zonooz, and Elahe Arani.
\newblock Trire: A multi-mechanism learning paradigm for continual knowledge
  retention and promotion.
\newblock \emph{Advances in Neural Information Processing Systems}, 36, 2024.

\bibitem[Vitter(1985)]{vitter1985random}
Jeffrey~S Vitter.
\newblock Random sampling with a reservoir.
\newblock \emph{ACM Transactions on Mathematical Software (TOMS)}, 11\penalty0
  (1):\penalty0 37--57, 1985.

\bibitem[Wu et~al.(2016)Wu, Shen, Liu, Dick, and Van Den~Hengel]{wu2016value}
Qi~Wu, Chunhua Shen, Lingqiao Liu, Anthony Dick, and Anton Van Den~Hengel.
\newblock What value do explicit high level concepts have in vision to language
  problems?
\newblock In \emph{Proceedings of the IEEE conference on computer vision and
  pattern recognition}, pp.\  203--212, 2016.

\bibitem[Wu et~al.(2017)Wu, Teney, Wang, Shen, Dick, and Van
  Den~Hengel]{wu2017visual}
Qi~Wu, Damien Teney, Peng Wang, Chunhua Shen, Anthony Dick, and Anton Van
  Den~Hengel.
\newblock Visual question answering: A survey of methods and datasets.
\newblock \emph{Computer Vision and Image Understanding}, 163:\penalty0 21--40,
  2017.

\bibitem[Wu et~al.(2019)Wu, Chen, Wang, Ye, Liu, Guo, and Fu]{wu2019large}
Yue Wu, Yinpeng Chen, Lijuan Wang, Yuancheng Ye, Zicheng Liu, Yandong Guo, and
  Yun Fu.
\newblock Large scale incremental learning.
\newblock In \emph{Proceedings of the IEEE/CVF Conference on Computer Vision
  and Pattern Recognition}, pp.\  374--382, 2019.

\bibitem[Yoon et~al.(2018)Yoon, Yang, Lee, and Hwang]{yoon2018lifelong}
Jaehong Yoon, Eunho Yang, Jeongtae Lee, and Sung~Ju Hwang.
\newblock Lifelong learning with dynamically expandable networks.
\newblock In \emph{International Conference on Learning Representations}, 2018.

\bibitem[Zenke et~al.(2017)Zenke, Poole, and Ganguli]{zenke2017continual}
Friedemann Zenke, Ben Poole, and Surya Ganguli.
\newblock Continual learning through synaptic intelligence.
\newblock \emph{Proceedings of machine learning research}, 70:\penalty0 3987,
  2017.

\bibitem[Zhang et~al.(2022)Zhang, Zhao, Tsiligkaridis, and
  Zitnik]{zhang2022self}
Xiang Zhang, Ziyuan Zhao, Theodoros Tsiligkaridis, and Marinka Zitnik.
\newblock Self-supervised contrastive pre-training for time series via
  time-frequency consistency.
\newblock \emph{Advances in Neural Information Processing Systems},
  35:\penalty0 3988--4003, 2022.

\bibitem[Zhao et~al.(2018)Zhao, Gan, Rouditchenko, Vondrick, McDermott, and
  Torralba]{zhao2018sound}
Hang Zhao, Chuang Gan, Andrew Rouditchenko, Carl Vondrick, Josh McDermott, and
  Antonio Torralba.
\newblock The sound of pixels.
\newblock In \emph{Proceedings of the European conference on computer vision
  (ECCV)}, pp.\  570--586, 2018.

\end{thebibliography}
\bibliographystyle{collas2024_conference}

\newpage

\appendix
\section{Appendix}

\subsection{Hyperparameters}
\label{sec:hyper}
As the goal of our study was to understand the role and benefits of utilizing multiple modalities in CL and not to extract the best possible results and establish a state-of-the-art, we did not conduct an extensive hyperparameter search for different buffer sizes and settings. For all our experiments, we set $\lambda$ to 0.01. For Dom-VGGSound, we use $\beta$=0.1 and for Seq-VGGSound and GCIL-VGGSound, we used $\beta$=1. Note that the results can be improved with hyperparameter tuning. 

\subsection{MMCL Benchmark Details}
\label{sec:VGG}
We designed the \textit{multimodal Continual Learning (MMCL)} benchmark with the following principles: (1) Adherence to the desiredata's in \cite{farquhar2018towards};
(2) Correspondence with unimodal benchmarks;
(3) Assessment of challenges that a learning agent encounters in the real world; and
(4) Accessibility to the wider research community.

Therefore, we considered the challenging Class-IL, Domain-IL, and Generalized-Class-IL (GCIL) scenarios, each of which simulates different sets of challenges for a learning agent in our dynamic and complex environment. To make the benchmark accessible to the wider research community and to facilitate further development of multimodal CL methods, we kept the overall dataset size and the number of classes similar to the widely adopted CIFAR-100 dataset. This ensures that the benchmark does not require excessively intensive computational and memory resources for experimentation.

\begin{figure}[th]
    \centering
    \includegraphics[trim=0 0 0.5cm 0, clip, width=1\linewidth]{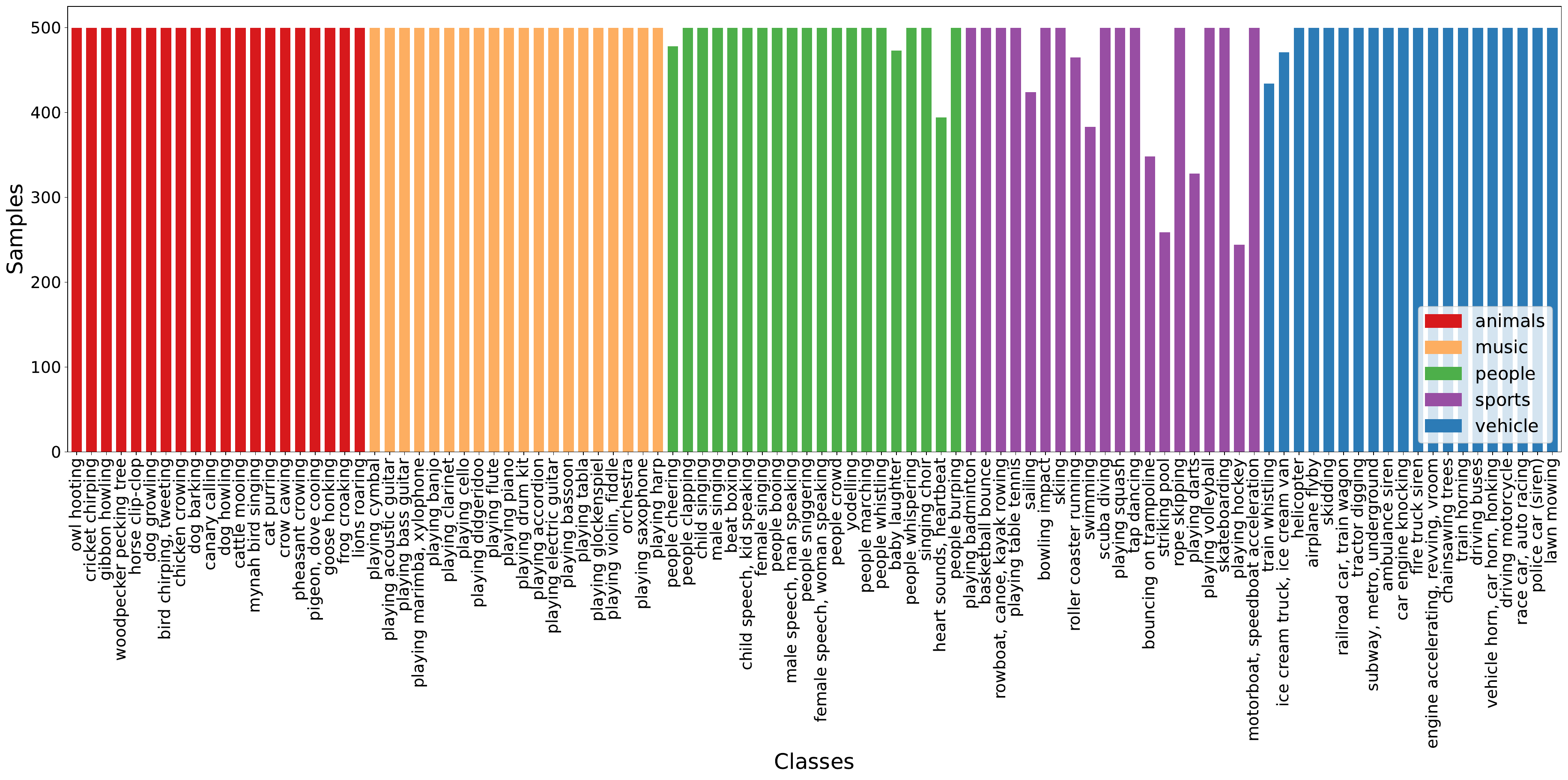}
    \caption{The distribution of training samples count for the subclasses in each supercategory in the MMCL benchmark. We consider 5 supercategories with 20 subclasses each and aim for a uniform sample count of 500 training samples and 50 test samples.
    }
    \label{fig:counts}
\end{figure}

We selected a subset of 5 supercategories (animals, music, people, sports, and vehicles) from the VGG-Sound dataset, each containing 20 subclasses (see Figure \ref{fig:counts}). This resulted in an overall class count of 100, similar to CIFAR-100. As with CIFAR-100, we aimed to have 500 training samples and 50 test samples for each class. However, due to the distribution of classes in the original VGG-Sound dataset and the current unavailability of some YouTube videos, it was not possible to acquire 500 samples for all the classes. Nevertheless, for the vast majority of classes, our benchmark is based on a uniform set of 500 training samples (see Figure \ref{fig:counts}). Notably, for the `sports' supercategory, we have a lower number of samples for the classes.

Figure \ref{fig:counts} provides details about the selected subclasses within each of the 5 supercategories and their respective training sample counts. For \textbf{Seq-VGGSound}, we randomly shuffled the classes and divided the dataset into 10 disjoint tasks, each containing 10 classes (the order of classes is each supercategory is provided in Figure \ref{fig:counts}). In \textbf{Dom-VGGSound}, we assigned the supercategory as the target label and created 10 tasks, with each task consisting of samples from the next two subclasses in the order presented in Figure \ref{fig:counts}. For example, in Task 1, we utilized samples from the `owl hooting' and `cricket chirping' subclasses for the `animals' supercategory, and in Task 2, we used `gibbon howling' and `woodpecker pecking tree,' and so forth. In the case of \textbf{GCIL-VGGSound}, we followed settings similar to \cite{mi2020generalized}: 20 tasks, each with a total of 1000 samples, and a maximum of 50 classes in each task. For reproducibility, we fixed the GCIL seed at 1992, given the probabilistic nature of GCIL. For further reproducibility and to make the dataset available to the research community at large, the dataset files and code for the method and benchmark will be made publicly available upon acceptance.

\begin{figure}[t]
    \centering
    \includegraphics[ width=.95\linewidth]{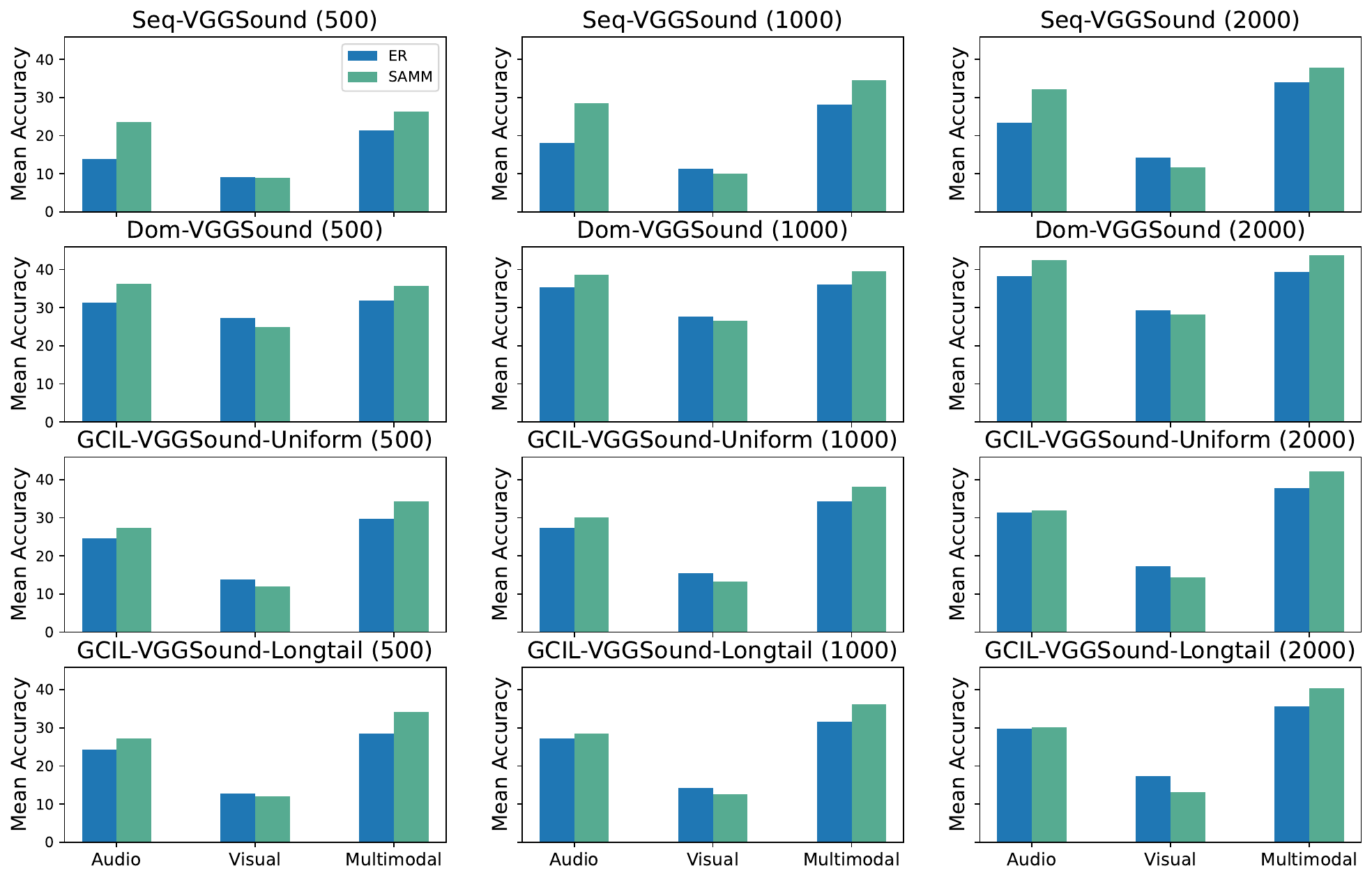}
    \caption{Visualization of the comparison of different methods on individual and multiple modalities on different CL scenarios based on the VGGSound dataset.}
    \label{fig:results-viz}
\end{figure}

\subsection{Additional Results}
To ensure that the performance improvements are not coming from the additional parameters in the MultiModal setup compared to the Unimodal models, we ran a control experiment where we reduce the number of channels to half in both the visual and audio encoder. This results in a multimodal architecture with roughly the same number of parameters as the unimodal counterparts. Table \ref{tab:vgg-sound-additional} shows that it only results in a slight decrease in performance, and it is still considerably better at learning and mitigating forgetting than unimodal setting with roughly the same number of parameters. This shows that performance improvements over the unimodal model cannot be attributed to the additional parameters in multimodal setting. 

Furthermore, to simulate the real-world scenario where often two modalities are captured with different sampling frequencies, we change the sampling frequency of the audio modality and measure the effect on performance. Note that we used a sampling rate of 16000 Hz for all of the experiments. To calculate the effect of different sampling frequency, we train the models with 10000 Hz and 20000 Hz with 1000 buffer size. Table \ref{tab:vgg-sound-additional} that the performance of SAMM on audio and multimodal data generally increases as the audio sampling frequency increases. However, for all the considered sampling frequencies, SAMM is able to utilize the complementary information across the modalities, and enhances the performance of the model compared to the unimodal settings. The results suggest that SAMM is able to utilize the complementary information across the modalities at different sampling frequencies. 

\begin{table}[t]
\caption{Additional experiments on Seq-VGGSound}
\label{tab:vgg-sound-additional}
\centering
\begin{tabular}{@{\extracolsep{4pt}}ll|ccc}
\toprule
Buffer & Method & Audio & Visual &  Multimodal \\ \midrule
 \multirow{6}{*}{1000} & SAMM (Original) & 28.59\tiny±0.77 & 10.08\tiny±0.34  & 34.51\tiny±2.37  \\ \cmidrule{2-5}
 &  ER   & 18.06\tiny±0.44 & 11.23\tiny±0.57 & 28.09\tiny±0.77  \\
 & SAMM (0.5$\times$ model size) & 26.65 & 9.76 & 32.93 \\ \cmidrule{2-5} 
 & SAMM (SR=10000)       & 24.61 & 8.92 & 33.82 \\
 & SAMM (SR=20000)       & 29.07 & 9.51 & 35.08 \\ \bottomrule
\end{tabular}
\end{table}

\subsection{Additional Analysis}
\begin{figure}[t]
    \centering
    \includegraphics[ width=.95\linewidth]{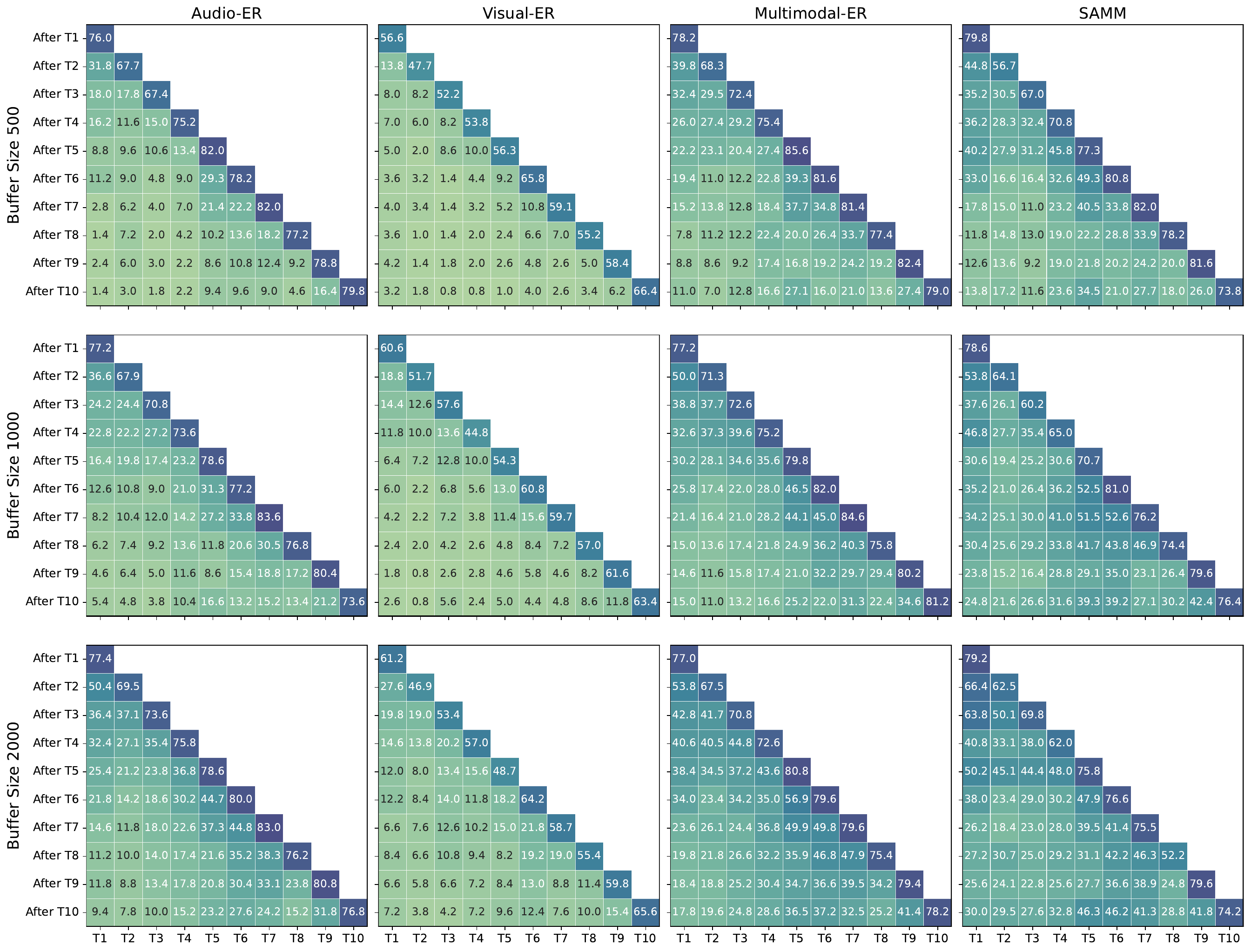}
    \caption{Taskwise performance of models trained with experience replay with different buffer sizes on multimodal vs. unimodal (audio and visual) data on Seq-VGGSound. As we train on new tasks, T (y-axis), we monitor the performance on earlier trained tasks (x-axis).}
    \label{fig:taskwise-all-seq-vgg}
\end{figure}

\begin{figure}[t]
    \centering
    \includegraphics[ width=.95\linewidth]{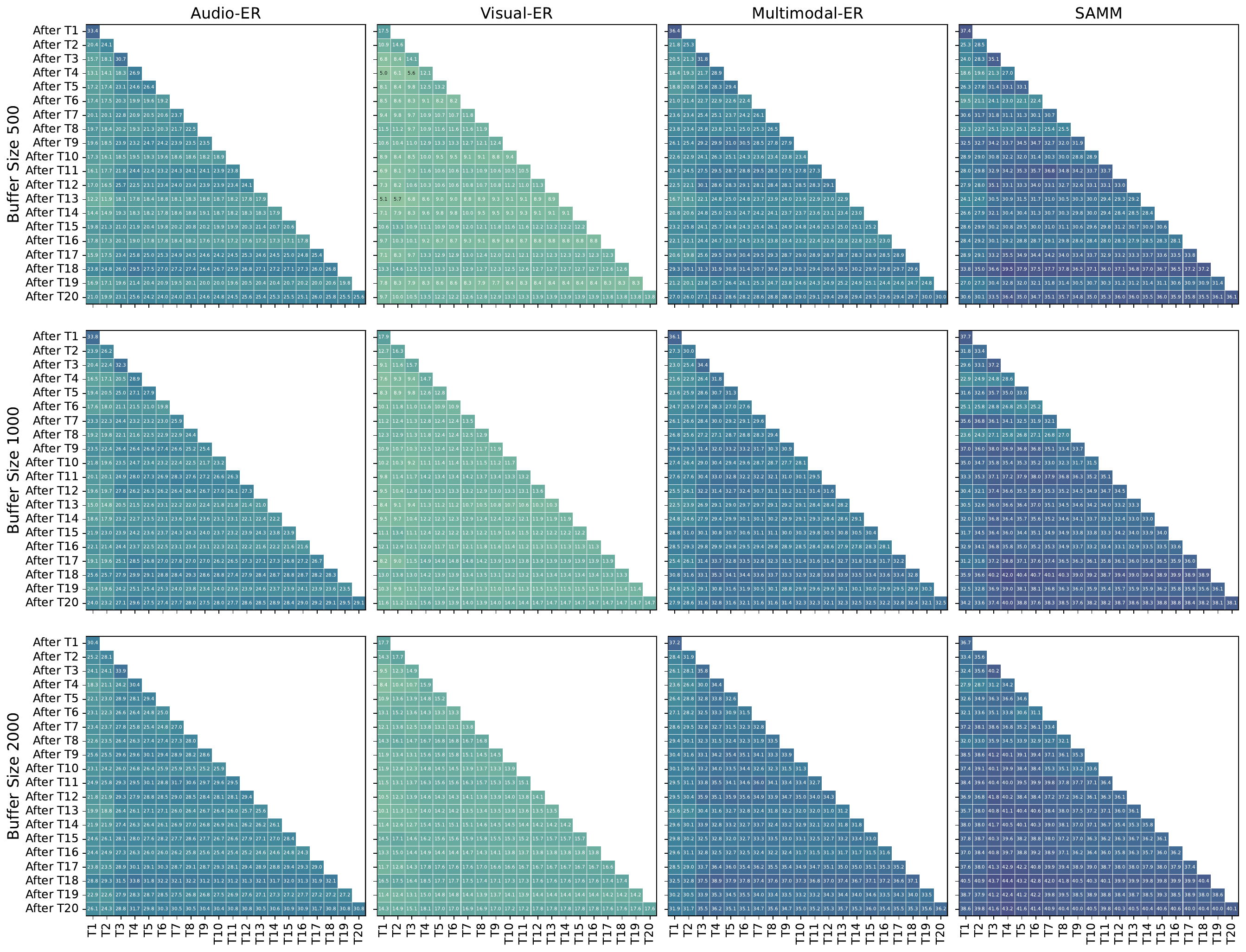}
    \caption{Taskwise performance of models trained with experience replay with different buffer sizes on multimodal vs. unimodal (audio and visual) data on GCIL-VGGSound (Longtail). As we train on new tasks, T (y-axis), we monitor the performance on earlier trained tasks (x-axis).}
    \label{fig:taskwise-all-gcil}
\end{figure}

\begin{figure}[t]
    \centering
    \includegraphics[ width=.95\linewidth]{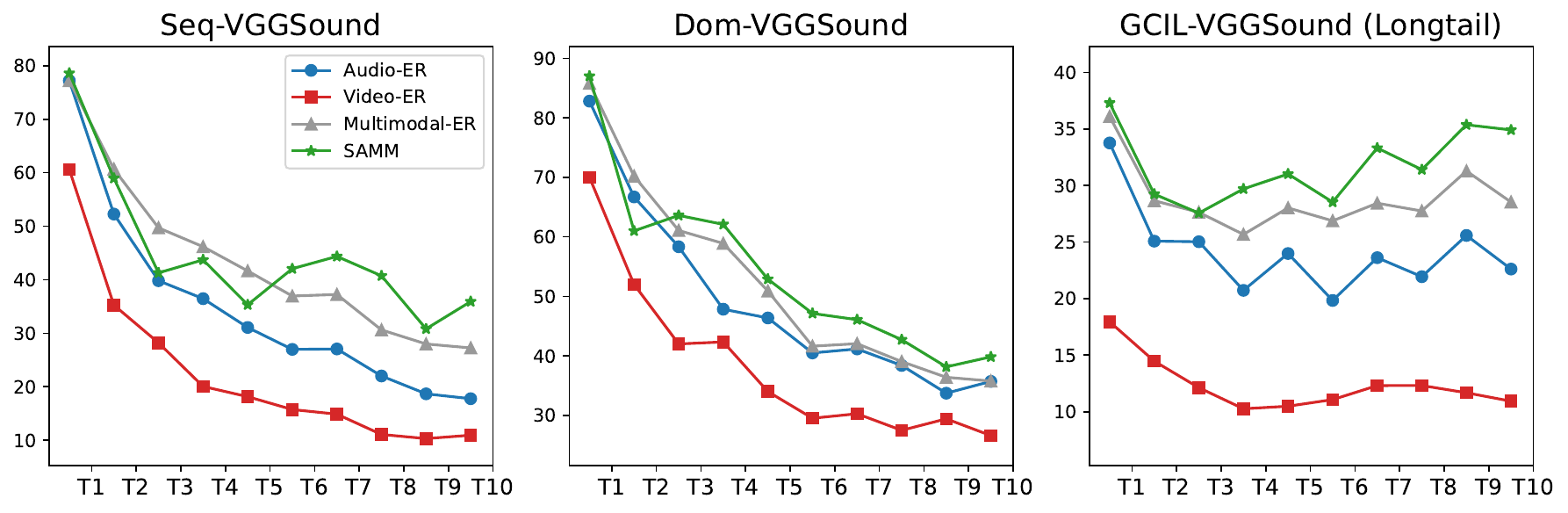}
    \caption{Mean accuracy of the different models on all the tasks (T) seen as training progresses.}
    \label{fig:mean-acc}
\end{figure}

\begin{figure}[t]
    \centering
    \includegraphics[ width=.95\linewidth]{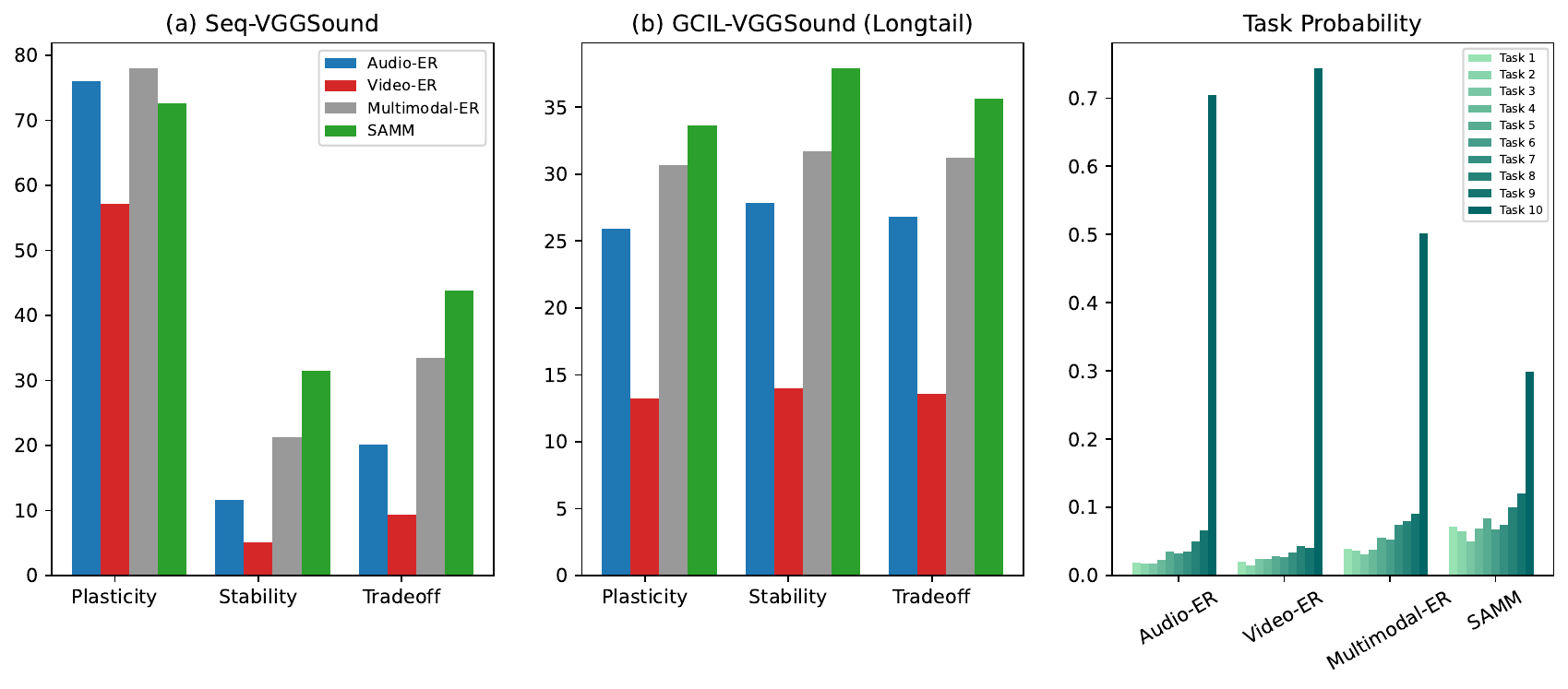}
    \caption{Comparison of Plasticity Stability tradeoff on (a) Seq-VGGSound and (b) GCIL-VGGSound (Longtail) with 1000 buffer size. (c) Comparison of the task recency bias. MMCL provides a better tradeoff between plasticity and stability and reduces the task recency bias on Seq-VGGSound with 1000 buffer size.}
    \label{fig:combined_appendix}
\end{figure}

Here, we extend the analysis performed in Section \ref{analysis-case}. Figure \ref{fig:taskwise-all-seq-vgg} shows the taskwise performance of the models as they are trained on Seq-VGGSound with different buffer sizes. Figure \ref{fig:taskwise-all-gcil} provides the taskwise performance on GCIL-VGGSound (Longtail). SAMM consistently improves knowledge retention across different datasets and buffer regimes without impeding the ability of the model to learn the new task. This is further supported by the mean accuracy plots after each task (Figure \ref{fig:mean-acc}) across all the datasets. Figure \ref{fig:combined_appendix} shows that SAMM provides a better trade-off between the stability and plasticity of the model and significantly reduces the task recency bias. Table \ref{tab:stab-plasticity} provides the mean stability plasticity statistics over 3 seeds on Seq-VGGSound with 1000 buffer size. SAMM provides a considerably better trade-off.

\begin{table}[h]
\centering
\caption{Stability Plasticity Tradeoffs for the different methods on Seq-VGGSound with 2000 buffer size. We report the mean and std over 3 seeds.}
\label{tab:stab-plasticity}
\begin{tabular}{l|ccc}
\toprule
Model           & ~~Plasticity~~  & ~~Stability~~   & ~~Tradeoff~~    \\ \midrule
Audio-ER        & 75.51\tiny±0.46 & 11.41\tiny±0.16 & 19.82\tiny±0.25  \\ 
Video-ER        & 57.15\tiny±0.05 & ~5.06\tiny±0.05 & ~9.29\tiny±0.09  \\ 
Multimodal-ER   & 77.87\tiny±0.12 & 18.60\tiny±2.66 & 29.94\tiny±3.74  \\ \midrule
SAMM            & 72.13\tiny±0.49 & 28.74\tiny±2.68 & \textbf{41.05}\tiny±2.82  \\ \bottomrule
\end{tabular}
\end{table}

\subsection{Summary of Contributions}
The primary objective of our study was to investigate the role of multiple modalities in facilitating continual learning (CL) and advocate for the shift towards multimodal CL. Our study is based on the premise that the human brain's capacity for processing multisensory information is a pivotal factor contributing to its continual learning capabilities. Our empirical findings present a compelling case for the transition to MultiModal CL. The results underscore that leveraging multiple modalities enables the model to learn more holistic and robust representations of objects and actions. Moreover, our study suggests that employing multimodal replay improves knowledge consolidation, and mitigates forgetting.

However, to fully explore the potential and benefits of the increasing amount of multimodal data in real-world applications in enhancing the lifelong learning capability of DNNs, for enhancing the lifelong learning capability of DNNs, it is imperative to extend the traditional unimodal CL benchmarks to encompass multimodal scenarios. To this end, we introduced a standardized Multimodal Continual Learning benchmark, which aims to simulate challenging and realistic real-world CL scenarios while maintaining correspondence with unimodal CL benchmarks and scenarios. We design the benchmark in a manner to makes it computationally feasible for the research community at large and test critical components of an efficient CL agent. It also allows the community to compare the benefits of multimodal CL over the traditional unimodal CL by drawing direct correspondence. We used visual and audio cues as it is the natural way that we interact with objects in the real world and these two modalities have correspondence and can be sampled in a fairly uniform manner. 

Our primary contribution lies in demonstrating the performance advantages of multimodal learning and experience replay over unimodal approaches in challenging CL scenarios, mirroring established CL settings in unimodal contexts. Furthermore, we propose promising avenues for aligning modalities to enhance complementary learning, furnishing a robust foundation and guiding principles for subsequent research endeavors. 

The findings from our study can be extended to incorporate additional modalities, such as language. By integrating language as an additional modality into the MMCL framework, the synergistic effects of multimodal learning across different modalities can be explored. As our study demonstrates the benefits of leveraging multiple sensory modalities for enhanced object and action representation, integrating language can provide further context and semantic understanding to the learning process. This extension would enable DNNs to learn more holistic and robust representations by fusing information from multiple modalities, leading to increased lifelong learning capabilities. Additionally, by including language as a modality within the MMCL benchmark, more complex and realistic CL scenarios can be created that better reflect the multifaceted nature of real-world learning environments.

Our study calls upon the research community to recognize the significance of multimodal CL, highlighting its potential as a pivotal component in achieving effective CL in DNNs. We advocate for a collective transition from unimodal CL to multimodal CL, drawing inspiration from parallels in human learning processes.

\end{document}